\pdfoutput=1
\documentclass[11pt]{article}
\usepackage{acl}
\usepackage{custom}

\title{Improving Minimum Bayes Risk Decoding with Multi-Prompt}

\author{
David Heineman, Yao Dou, Wei Xu \\
School of Interactive Computing, Georgia Institute of Technology \\
{\small \texttt{\{david.heineman, douy\}@gatech.edu; wei.xu@cc.gatech.edu}} \\
}

\begin{document}
\selectlanguage{english}
\maketitle

\begin{abstract}
While instruction fine-tuned LLMs are effective text generators, sensitivity to prompt construction makes performance unstable and sub-optimal in practice. 
Relying on a single `best' prompt cannot capture all differing approaches to a generation problem.
Using this observation, we propose \textit{multi-prompt decoding}, where many candidate generations are decoded from a prompt bank at inference-time. 
To ensemble candidates, we use Minimum Bayes Risk (MBR) decoding, 
which selects a final output using a trained value metric.
We show multi-prompt improves MBR across a comprehensive set of conditional generation tasks (Figure \ref{fig:multi_prompt_small}), and show this is a result of estimating a more diverse and higher quality candidate space than that of a single prompt.
Further experiments confirm multi-prompt improves generation across tasks, models and metrics.\footnote{Our experiment code, data and prompts are available at \url{https://github.com/davidheineman/multi-prompt}.}
\end{abstract}

\section{Introduction}
\label{sec:introduction}
\begin{figure}[t!]
\centering
\includegraphics[width=0.44\textwidth]{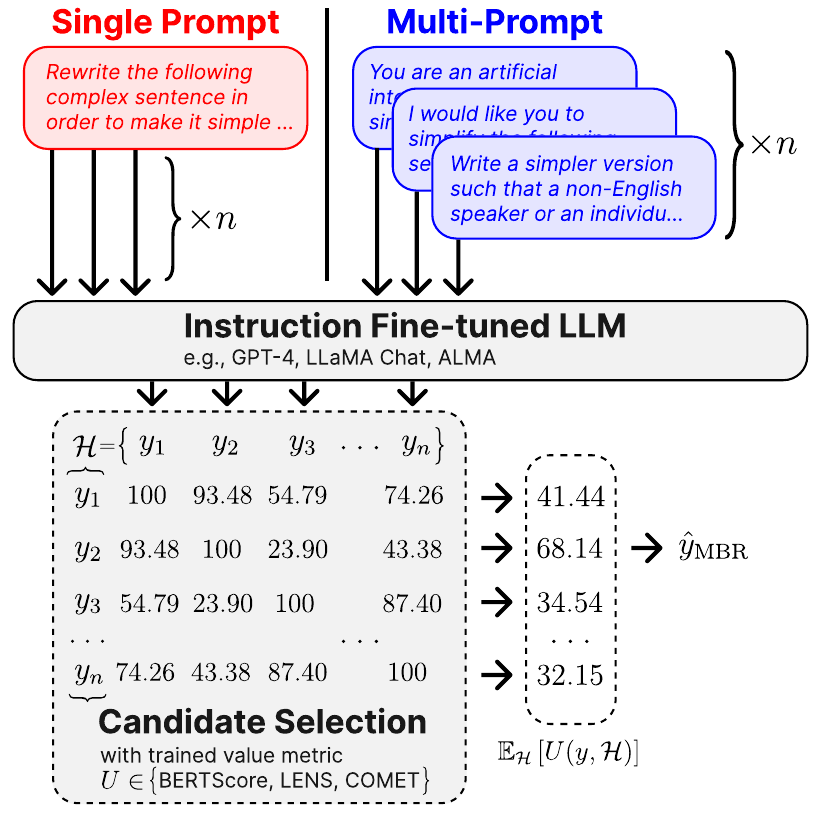}
\setlength{\belowcaptionskip}{-7pt}
\captionof{figure}{Multi-prompt MBR generates candidates using a human- or model-written prompt bank and selects the highest pairwise score with a trained value metric.}
\label{fig:overview}
\end{figure}

Minimum Bayes Risk (MBR) decoding \cite{bickel1977mathematical} improves the generation quality of large language models (LLMs) over standard, single-output decoding methods, such as beam search and sampling. 
MBR generates a set of candidates and selects the one with the highest expected utility, using all other hypotheses as references (see Fig. \ref{fig:overview}, left), following a simple intuition that a desirable output should be highly probable and consistent with others. 
MBR has been applied across a variety of NLP generation tasks \cite{amrhein-sennrich-2022-identifying,shi-etal-2022-natural,suzgun-etal-2023-follow,jain2023self}. In particular, self-consistency \cite{wang2023selfconsistency}, a special case of MBR, has become widely used to improve LLM reasoning capabilities by ensembling reasoning paths.

\begin{figure}[t!]
\centering
\includegraphics[width=0.49\textwidth]{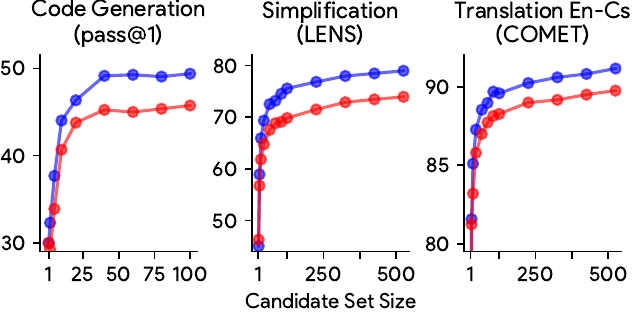}
\setlength{\abovecaptionskip}{-5pt}
\setlength{\belowcaptionskip}{-14pt}
\captionof{figure}{\textcolor{blue}{Multi-prompt} and \textcolor{red}{single prompt} MBR results for code generation on \textsc{HumanEval}, text simplification on \textsc{SimpEval}, and translation on WMT '22 \textsc{En-Cs} generated with open-source 7B LLMs (details in \S\ref{sec:experiments}).}
\label{fig:multi_prompt_small}
\end{figure}

A central question to improve the generation quality of MBR decoding is how to balance between diversity and adequacy within the candidate set.
Prior work has found success using sampling-based decoding to generate diverse hypotheses  \cite{eikema-aziz-2020-map,freitag-etal-2022-high, freitag-etal-2023-epsilon}. However, naively increasing the sampling temperature eventually degrades the quality of the candidates.
Recently, instruction fine-tuned LLMs \cite{ouyang2022training,chung2022scaling} have opened up the possibility of writing \textit{prompts} in various formats to elicit higher diversity generations.
As these models are observed to be sensitive to prompt design, a slight change in phrasing or the inclusion of more relevant example can significantly impact model behavior \citep{srivastava2023beyond,white2023prompt}.

Taking advantage of the prompt sensitivity of LLMs, we introduce multi-prompt MBR decoding, which samples candidates using a bank of human- or model-written prompts (see Figure \ref{fig:overview}, right). 
Intuitively, exploring a variety of prompts enables the generation of diverse, high quality hypotheses that provide a closer representation of the true output distribution.
By guiding the model towards different regions of the output space, each prompt captures unique sequences that are coherent and relevant to the specific input example.

We experiment with three distinct generation tasks: text simplification \cite{maddela-etal-2023-lens}, machine translation \cite{kocmi-etal-2022-findings}, and code generation \cite{chen2021evaluating}.
Each task assess the impact of different prompt components on multi-prompt MBR, such as instance-level prompts for code, task descriptions for simplification, and in-context examples for translation.
To account for the relative quality between prompts, we develop different strategies for selecting prompts that outperform a baseline random choice: \textit{sampling} prompts from a large prompt bank based on their usage on an unlabeled set of task data and \textit{selecting} prompts using embedding-based heuristics without any examples.

We evaluate multi-prompt MBR on a broad range of LLMs including open-source models such as Llama 2 \cite{touvron2023llama} and state-of-the-art closed-source models such as GPT-4 \cite{achiam2023gpt}. Our results show multi-prompt MBR consistently improves single-prompt MBR across all three tasks and model scales, with gains of up to 7\% on HumanEval \cite{chen2021evaluating} and 5 points of LENS score on \textsc{SimpEval} \cite{maddela-etal-2023-lens}. Figure \ref{fig:multi_prompt_small} displays results for models at the 7B scale.
Finally, we study the dynamics between different utility and evaluation metrics, revealing that multi-prompt MBR with one metric improves performance universally across metrics.

\section{Preliminaries}
\label{sec:preliminaries}
Instruction fine-tuned LLMs are trained to follow arbitrary natural language task descriptions \citep{wei2022finetuned}.
Given an input $x$ and prompt $\rho$, an autoregressive language model $\pi_\theta$ parameterized by $\theta$ estimates an output sequence $y \sim \pi_\theta(x, \rho)$ using an decoding algorithm 
by sampling the next token conditioned on the input $\pi_\theta(y_i|y_{<i}, x, \rho)$.
The decoding algorithm aims to generate $y$ by maximizing the sequence likelihood over the language model distribution $\pi_\theta(y|x, \rho)=\Pi_{i=1}^T \pi_\theta(y_i|y_{<i}, x, \rho)$. 

\begin{figure*}[t!]
\centering

\adjustbox{valign=b}{
    \subfloat[\label{fig:candidate-probabilities-a}]{
        \includegraphics[width=0.68\linewidth]{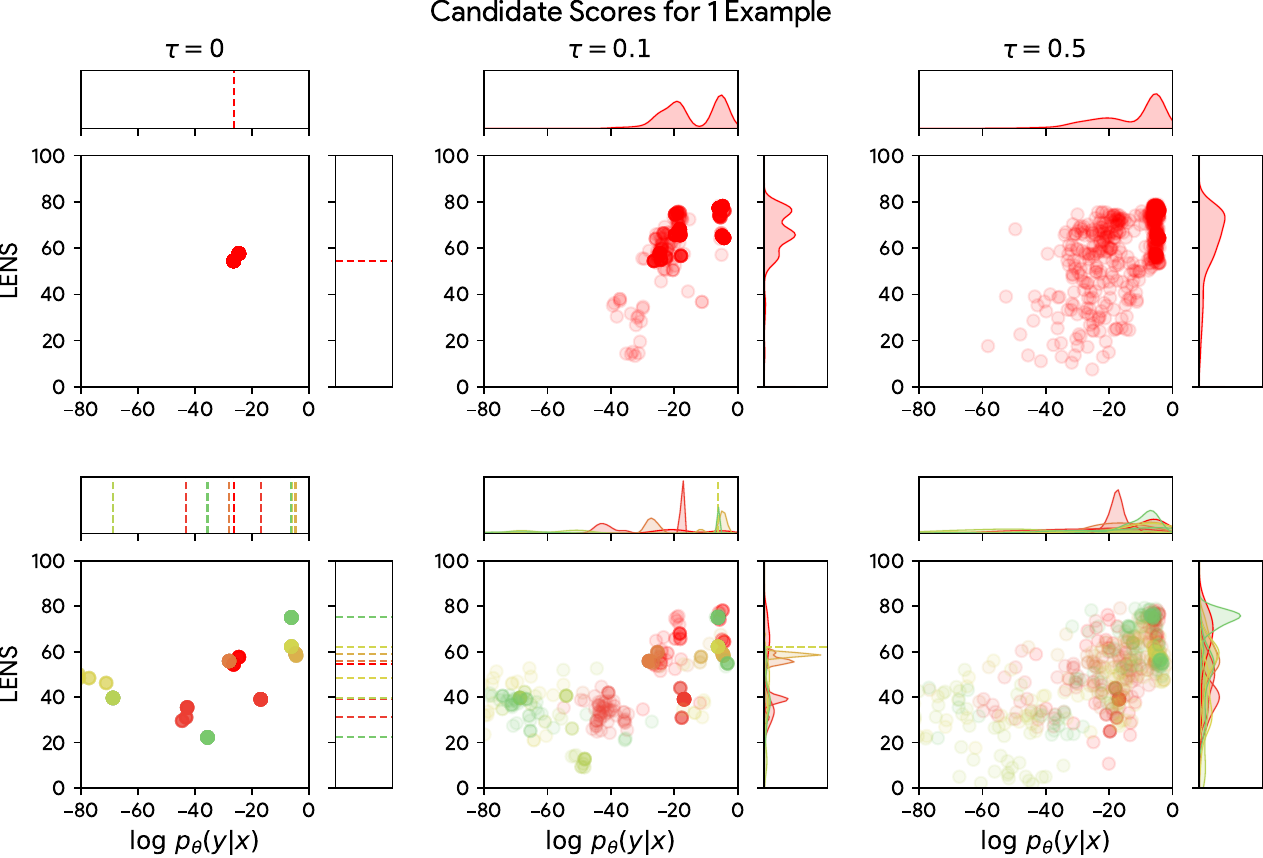}
    }
}
\adjustbox{valign=b}{
    \begin{tabular}{@{}c@{}}
        \subfloat[\label{fig:candidate-probabilities-b}]{
            \includegraphics[width=0.25\linewidth]{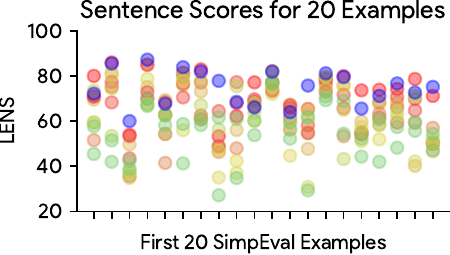}
        } \\
        \subfloat[\label{fig:candidate-probabilities-c}]{
            \includegraphics[width=0.25\linewidth]{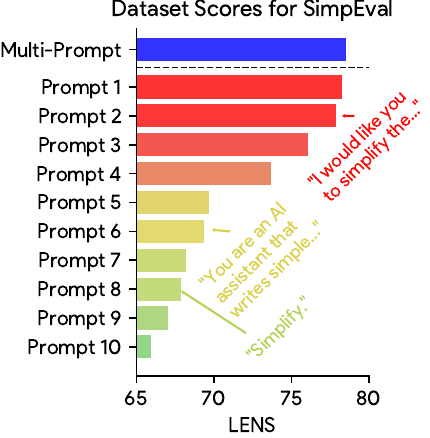}
        }
    \end{tabular}
}

\setlength{\abovecaptionskip}{7pt}
\setlength{\belowcaptionskip}{-7pt}
\captionsetup{font=small}
\captionof{figure}{
(a) \textsc{Lens} score and sequence probability for 1000 generations on a single text simplification example decoded from Llama 2 7B Chat with temperatures $\tau=\left[0, 0.1, 0.5\right]$ using a single prompt (top) and multiple prompts (bottom). As the temperature increases, we find each prompt estimates candidate sequences centered at different modes. 
(b) \textsc{Lens} scores of the best generation per-prompt for the first 20 sentences in \textsc{SimpEval}, showing no single prompt produces the best overall output.
(c) Dataset-level LENS performance of each prompt when performing single prompt MBR vs. multi-prompt MBR.
}
\label{fig:candidate-probabilities}
\end{figure*}

\vspace{2pt}
\pg{Minimum Bayes Risk Decoding.}
In practice, the highest likelihood sequence does not necessarily yield the highest quality generation \citep{jaeger2006speakers}. % jaeger2006speakers, meister-etal-2022-high, freitag-etal-2022-high
From this observation,
MBR decoding \cite{bickel1977mathematical,eikema-aziz-2020-map} first samples a set of hypotheses $\mathcal{H}$ from the model $\pi_\theta$, approximating the true distribution of output space $\mathcal{Y}$, then selects the output $\hat{y}_{MBR}$ that maximizes the expected utility (or minimizes the expected loss in traditional formulation) with respect to a set of references $\mathcal{R}$:
\begin{equation}
\hat{y}_{\text{MBR}}=\argmax_{y\in \mathcal{H}} \left( \mathbb{E}_{\mathcal{H}\sim \pi_\theta} [\,U(y,\mathcal{R})] \right),
\end{equation}

\noindent
where $U(y, R)=\mathbb{E}_{y'\sim \mathcal{R}} [u(y, y')]$ and $u(y, y')$ is a utility function that evaluates hypothesis $y$ against a reference $y'$. In practice, $\mathcal{R}$ is also sampled from the same model $\pi_\theta$ under the assumption that the model produces reliable outputs in expectation, and is usually set as identical to hypothesis set $\mathcal{H}$.

Many existing techniques to improve LLMs' performance such as self-consistency \citep{wang2023selfconsistency} and output ensemble \cite{kobayashi-2018-frustratingly} are special cases of MBR. For instance, self-consistency can be viewed as MBR using the utility function $u(y, y')=\mathds{1}\left[\text{ans}(y) = \text{ans}(y')\right]$, where $\text{ans}(y)$ is the answer extracted from the reasoning path $y$ \citep{bertsch-etal-2023-mbr}.

\section{Multi-Prompt MBR Decoding}
\label{sec:method}
Prior work on MBR decoding primarily uses models trained or fine-tuned for a specific generation task \cite{freitag-etal-2022-high, fernandes-etal-2022-quality}. 
With instruction fine-tuned LLMs, the input $x$ is contained within a structured prompt $\rho$, consisting of task instruction and/or in-context examples. Earlier studies have extensively documented that the design of the prompt has a dramatic impact on overall performance \cite{mishra-etal-2022-reframing, khashabi-etal-2022-prompt, lu-etal-2022-fantastically, sclar2023quantifying}.

To investigate this phenomenon, we show in Figure \ref{fig:candidate-probabilities-a} (bottom) the likelihoods and quality of samples from 10 prompts of varying performance for a text simplification task, measuring quality as the \textsc{Lens} metric score against a set of gold references. 
Greedy sampling ($\tau\!=\!0$) estimates different sequences for each instruction, with single prompt (Figure \ref{fig:candidate-probabilities-a}, top) generating a single sequence.
As we increase temperature $\tau$, generations from a single prompt simply exhibit noise centered around the mode of the highest likelihood sequence, while multi-prompt estimates a generations around modes uniquely defined by each prompt.
For instance, one of the prompts (i.e., Prompt 9 highlighted in green) produces the highest quality generation for this one input sentence, despite having a low performance over the entire dataset. In fact, no prompt consistently produces the highest quality sequences, as illustrated in Figure \ref{fig:candidate-probabilities-b}, rather prompts are most effective at different inputs. 

Building upon these insights, we propose multi-prompt MBR decoding, depicted in Figure \ref{fig:overview}, where the MBR hypothesis set $\mathcal{H}$ consists of outputs sampled from $n$ distinct prompts $\rho$:
\begin{equation}
\mathcal{H} = \bigcup_{i=1}^n \mathcal{H}_i,\:\text{where}\: \mathcal{H}_i = \{y | y \sim \pi_\theta(x, \rho_i)\}.
\end{equation}
\citet{bertsch-etal-2023-mbr} show that MBR seeks the mode of some distribution $q$ over a quality feature $\phi(y)$ applied to the output space rather than the mode of the model's distribution:
\begin{equation}
    \hat{y}_{\text{MBR}} \approx \argmax_{y\in \mathcal{H}} q(\phi(y)|x).
\end{equation}
We hypothesize, in expectation, the mode of $\phi(y)$ across outputs from multiple prompts has higher downstream performance compared to that derived from a single prompt. 
This is empirically supported by our example, where Figure \ref{fig:candidate-probabilities-c} shows that multi-prompt MBR outperforms individual single-prompt MBR across the full task dataset.

Although multi-prompt ensembles hypothesis spaces between prompts, some notion of objective quality still exists when constructing the prompt bank. 
As shown in Figure \ref{fig:candidate-probabilities-c}, the majority of the 10 human-written prompts fall within a 10-point range of LENS scores when evaluated on the task dataset but a few prompts consistently produce low-quality generation.
Therefore, to account for the hierarchy in prompt quality, we propose two methods for choosing the prompts used at generation time from a prompt bank $\mathcal{P}$:
\uline{sampling} from a learned distribution of prompts, based on a small unlabeled train set (\S \ref{subsec:prompt_sampling}); and \uline{selecting} a subset of prompts based on heuristics in the absence of a train set (\S \ref{subsec:prompt_selection}).

\subsection{Prompt Sampling}
\label{subsec:prompt_sampling}
In this approach, we first calculate the probability of each prompt $p(\rho)$ as the proportion of times that prompt generates the highest scoring output on a separate training set.
At inference time, prompts are sampled with replacements from this learned probability distribution, and candidate outputs are then generated given these prompts.

\vspace{2pt}
\pg{Top-$p$ Prompt Sampling.} Inspired by the principle of nucleus sampling \cite{holtzman2020the}, our goal is to keep the prompts with high probability and truncate the least used prompts by setting their probabilities to zero.
We define the top-$p$ prompt set as the minimal set $\mathcal{P}_{\text{top-p}}\subseteq \mathcal{P}$ such that:
\begin{equation}
    \sum_{i=0}^{|\mathcal{P}_{\text{top-p}}|}{p(\rho_i)}\ge p.
\end{equation}
We then re-normalize the distribution of $\mathcal{P}_{\text{top-p}}$ and sample prompts from the new distribution:
\begin{equation}
    p'(\rho) = \begin{cases}
        \frac{p(\rho)}{\sum_{\rho \in \mathcal{P}_{\text{top-p}}} p(\rho)} & \text{if } \rho \in \mathcal{P}_{\text{top-p}} \\
        0 & \text{otherwise.}
    \end{cases}
\end{equation}

% \wei{explain what is the $p$} 

\subsection{Prompt Selection}
\label{subsec:prompt_selection}
Prompt selection chooses a fixed subset $\mathcal{P_{\text{best}}}\subset \mathcal{P}$ of $|\mathcal{P_{\text{best}}}|=k$ prompts based on heuristics. Compared to sampling, this does not require an additional training set to evaluate prompt efficacy.
We consider the following heuristics for selecting $\mathcal{P_{\text{best}}}$: prompts that have the \uline{closest similarity} and \uline{greatest dissimilarity} with others, and prompts that are randomly selected from each \uline{$k$-NN cluster}, which is also useful when a training set is presented, allowing the selection of high-performing prompts within each cluster.
% In our experiments, 
We calculate the semantic (dis)similarity of prompts based on SentenceBERT \citep{reimers-gurevych-2019-sentence} embeddings.

\section{Experiment Setup}
\label{sec:experiments}
In this section, we describe the experimental details for evaluating the efficacy of multi-prompt MBR decoding across tasks, prompt setups, models, and utility metrics, with results and analyses in \S\ref{sec:results}.  

\subsection{Tasks \& Datasets}
Unlike previous work applying MBR to a single generation task \cite{shi-etal-2022-natural,eikema-aziz-2022-sampling}, we deliberately select three unique tasks to demonstrate the universality of multi-prompt: text simplification with task-level instructions, code generation with example-level instructions, and machine translation with in-context examples.

\vspace{2pt}
\noindent\textbf{Code Generation.} 
We use HumanEval \citep{chen2021evaluating} benchmark, where models are tasked with generating a Python program given a description with unit tests. 
Since each example is a unique coding task, we generate a unique prompt bank for each input. 
Following \citet{zhang2023coder}, we reject empty, degenerate (e.g., \texttt{pass}, \texttt{return None}), or non-compiling programs before applying MBR.

\vspace{2pt}
\noindent\textbf{Text Simplification.} 
We use the \textsc{SimpEval$_{2022}$} test set \citep{maddela-etal-2023-lens}, containing complex sentences from Wikipedia, paired with human-written simplifications. 
The prompt bank is generated based on author-written examples (Table \ref{table:human_written_prompts}) and are used for the entire dataset.
\vspace{2pt}

\noindent\textbf{Machine Translation.} 
We intentionally choose the EN $\rightarrow$ CS language pair from the WMT 22 \cite{kocmi-etal-2022-findings} \texttt{newstest} corpus, ensuring its exclusion from the training data of recent translation LLMs or metrics \citep{xu2024contrastive}. 
Results on additional language pairs are in Appendix \ref{appdx:translation_results}.

\subsection{Constructing the Prompt Bank}
\cam{
For text simplification and code generation experiments, we first collect a small set of manually written seed prompts and construct the full prompt set by using GPT-4 Turbo to generate diverse paraphrases of the seed prompts. 
% This approach follows existing studies in prompt sensitivity \citep{mizrahi2023state, gonen-etal-2023-demystifying}.
% Model-written prompts are generated using GPT-4 Turbo. 
The authors manually write 10 seed prompts for text simplification (Table \ref{table:human_written_prompts}) and use the original \textsc{HumanEval} instruction from each example as the seed prompt for code generation. 
For translation experiments, we use randomly sampled in-context examples taken from previous WMT shared tasks as the prompt bank instead of generating translation instructions. 
In our preliminary experiments, we found translation LLM performance to be more sensitive to varying examples rather than translation instructions.
}

For multi-prompt experiments, we select from the prompt bank with top-$p$ prompt sampling (\S\ref{sec:prompt_sampling}) using $p\!=\!0.6$, where the prompt usage $p(\rho)$ is calculated using a held-out 20\% split of each dataset. \cam{For our single prompt baselines, we use a randomly selected prompt from the prompt bank.} Human-written prompts and prompt generation instructions are included in Appendix \ref{appdx:prompts}.

\subsection{Models}
Our main experiments are performed with Llama 2-7B Chat \citep{touvron2023llama} for simplification, ALMA-7B-R \citep{xu2024contrastive} for translation and CodeLLaMA-13B Instruct \citep{roziere2023code} for code generation, all fine-tuned to follow instructions. In \S\ref{sec:multi_model} we further explore a wide range of model architectures and sizes, including state-of-the-art and task-specific fine-tuned models. Unless otherwise specified, we generate the hypothesis set using nucleus sampling \cite{holtzman2020the} with $\tau\!=\!0.9,p\!=\!0.95$. We include a detailed review of all models in this work in Appendix \ref{appdx:models}.

\subsection{Utility Metrics \& Evaluation} % and Evaluation
Our core experiments use the trained \textsc{Lens} \citep{maddela-etal-2023-lens} for simplification and \textsc{Comet} \citep{rei-etal-2020-comet} for translation as the candidate selection metric.
For code generation, we use \textsc{MBR-Exec} \citep{shi-etal-2022-natural}, which executes each candidate program against a set of test cases, selecting the program with the highest agreement over all test cases' outputs. As in \citet{zhang2023coder}, we use the docstring examples as test cases for \textsc{MBR-Exec} and evaluate with pass@1. 
Given the growing body of work on metric development, we verify our multi-prompt results across a broad range of utility and evaluation metrics in \S\ref{sec:metric_eval}.

\section{Experiment Results}
\label{sec:results}
We compare multi-prompt decoding to traditional MBR (\S\ref{sec:multi_prompt_results}), ablate the prompt sampling mechanism (\S\ref{sec:prompt_sampling}), vary model architectures (\S\ref{sec:multi_model}), evaluate across utility metrics (\S\ref{sec:metric_eval}) and finally evaluate multi-prompt on efficient MBR alternatives (\S\ref{sec:rerank}).

\begin{figure}[t!]
\centering
\includegraphics[width=0.47\textwidth]{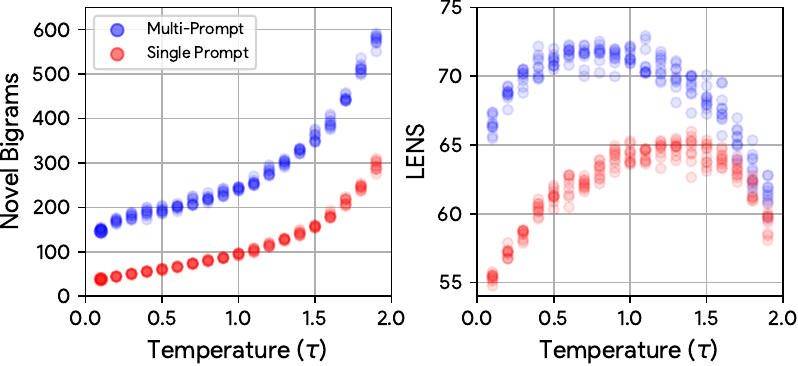}
\setlength{\abovecaptionskip}{4pt}
\setlength{\belowcaptionskip}{-10pt}
\captionof{figure}{Candidate set diversity and \textsc{Lens} scores on \textsc{SimpEval} for 200 repetitions of single-prompt and multi-prompt at various temperatures. At low temperatures, the increased candidate diversity from multi-prompt directly translates to improved performance.}
\label{fig:simplification_temperatures}
\end{figure}

\subsection{How does multi-prompt MBR perform?}
\label{sec:multi_prompt_results}

\pg{Multi-prompt Improves MBR.} 
We report our main results in Figure \ref{fig:multi_prompt_small}, which compares single prompt and multi-prompt performance when generating up to 500 candidates.
Multi-prompt consistently outperforms standard MBR for all tasks. 

\setlength{\tabcolsep}{2pt}
\begin{table}[t!]
\centering
\fontsize{8}{9.5}\selectfont
% \small

\begin{tabular}{p{100pt}cC{1pt}cC{1pt}c}

\toprule
& pass@1 && \textsc{Lens} && \textsc{Comet} \\
\midrule

\textit{Single Prompt} ($|\mathcal{H}|\!=\!100$)  & 48.78 && 74.67 && 88.93 \\
\hdashrule
\multicolumn{6}{l}{\textit{Multi-Prompt + Prompt Sampling} ($|\mathcal{P}|\!=\!100$)} \\

Random Selection        & -- && 74.91$^*$          && 89.98$^*$          \\
Prompt Sampling         & -- && 78.29$^*$          && 90.33$^*$          \\
Top-$p$ Prompt Random   & -- && 78.61$^*$          && 90.11$^*$          \\
Top-$p$ Prompt Sampling & -- && \textbf{79.08}$^*$ && \textbf{90.36}$^*$ \\

\midrule

\textit{Single Prompt} ($|\mathcal{H}|\!=\!10$)  & 41.55 && 61.26 && 87.24 \\
\hdashrule
\multicolumn{6}{l}{\textit{Multi-Prompt + Prompt Selection} ($\mathcal{P}_{\text{best}}\!\subset\! \mathcal{P}$, $|\mathcal{P}_{\text{best}}|\!=\!10$)} \\

Random Selection           & 39.63              && 60.00              && 87.81$^*$          \\
$k$-NN Cluster Random      & 40.24              && 58.73              && 87.80$^*$          \\
Farthest Similarity        & \textbf{44.51}$^*$ && 58.32              && \textbf{88.14}$^*$ \\
Closest Similarity         & 37.80              && 61.53$^*$          && 87.73$^*$          \\
Highest Performance        & --                 && 62.43$^*$          && 87.65              \\
$k$-NN Cluster Performance & --                 && \textbf{66.12}$^*$ && 87.73$^*$          \\

\bottomrule
\end{tabular}
\caption{Results for prompt sampling using 100 prompts (top) and subset selection using 10 of 100 prompts (bottom). \resp{* $=$ Statistically significant improvement with $p\!<\!0.05$.} Sampling from a weighted, truncated distribution improves multi-prompt across candidate set sizes.}
\label{table:prompt_selection}
\vspace{-6pt}
\end{table}

\vspace{2pt}
\pg{Candidate Diversity $\nRightarrow$ Quality.} 
To measure the impact of temperature on the candidate set quality, we report performance and diversity, as measured by novel bi-grams, across temperatures in Figure \ref{fig:simplification_temperatures}. 
For low temperatures, we find that multi-prompt generates a consistently more diverse candidate space, which directly translates to higher-quality generation. 
While single prompt MBR performance improves with temperature $\tau > 1$, despite generating an equal or greater diversity set than multi-prompt, multi-prompt MBR still produces higher quality candidates.
As $\tau \rightarrow 2$, the quality of single and multi-prompt MBR begins to degrade as their candidate sets become too noisy to generate high-quality sequences.
Framing the decoding process as each prompt estimating a unique distribution of candidate generations (\S\ref{sec:method}), the ability of multi-prompt to achieve higher quality generation as a result of candidate set diversity is intuitively the byproduct of combining multiple candidate distributions defined by each instruction. 

\begin{table}[t!]
\centering
% \small
\fontsize{8}{9.5}\selectfont

\begin{tabular}{L{60pt}C{30pt}C{30pt}C{40pt}C{40pt}}
\toprule
 & Single Prompt & Multi\nobreakdash-prompt & Cand. \textsc{Bleu} (MP on SP) & Cand. \textsc{Bleu} (SP on MP) \\
\midrule
\multicolumn{5}{l}{\textit{Code Generation} ($|\mathcal{H}|\!=\!20$) -- \textsc{HumanEval} (pass@1)} \\
\hdashrule
StarCoder 2 15B     & 44.51 & 49.39 & 49.69 & 50.13 \\ % (\plus{4.88})
CodeLlama 7B        & 37.80 & 40.85 & 62.05 & 63.32 \\ % (\plus{3.05})
CodeLlama 13B       & 43.29 & 48.17 & 59.49 & 60.76 \\ % (\plus{4.88})
CodeLlama 34B       & 45.73 & 52.44 & 61.59 & 62.92 \\ % (\plus{6.71})
CodeLlama 70B       & 61.59 & 68.90 & 63.15 & 65.12 \\ % (\plus{7.31})
GPT-3.5             & 68.29 & 73.78 & 83.07 & 89.86 \\ % (\plus{5.49})
GPT-4               & 81.71 & 82.93 & 81.72 & 89.82 \\ % (\plus{1.22})
\midrule
\multicolumn{5}{l}{\textit{Text Simplification} ($|\mathcal{H}|\!=\!100$) -- \textsc{SimpEval} (LENS)} \\
\hdashrule
Ctrl T5 3B          & 72.6 & -- & -- & -- \\
Ctrl T5 11B         & 74.4 & -- & -- & -- \\
Llama 2 7B Chat     & 75.71 & 80.38 & 80.71 & 74.68 \\ % (\plus{4.67})
Llama 2 13B Chat    & 78.19 & 80.27 & 79.30 & 77.65 \\ % (\plus{2.08})
Llama 2 70B Chat    & 82.21 & 83.28 & 74.11 & 70.65 \\ % (\plus{1.07})
GPT-3.5             & 76.87 & 81.25 & 94.18 & 85.56 \\ % (\plus{4.38})
GPT-4               & 76.47 & 81.56 & 96.74 & 81.05 \\ % (\plus{5.09})
\midrule
\multicolumn{5}{l}{\textit{Translation} ($|\mathcal{H}|\!=\!100$) -- WMT '22 \textsc{En-Cs} (COMET)} \\
\hdashrule
WMT '22 Winners     & 91.9 & -- & -- & -- \\
MS Translate API    & 90.6 & -- & -- & -- \\
ALMA 7B R           & 89.17 & 89.94 & 87.22 & 81.20 \\ % (\plus{0.77})
ALMA 13B R          & 89.41 & 90.45 & 89.75 & 84.74 \\ % (\plus{1.04})
GPT-3.5             & 91.27 & 91.35 & 99.26 & 95.47 \\ % (\plus{0.08})
GPT-4               & 92.24 & 92.47 & 90.21 & 90.85 \\ % (\plus{0.23})
% TowerInstruct 7B    & 86.69 & 88.58 \\
% TowerInstruct 13B   & 88.92 & 90.00 \\
% Aya 101             & 90.27 & 90.68 \\
\bottomrule
\end{tabular}
\setlength{\belowcaptionskip}{-10pt}
\caption{Metric scores for state-of-the-art systems compared to LLMs with multi-prompt using $|\mathcal{H}|$ candidates. Translation and simplification baselines are as reported in \citet{hendy2023good} and \citet{maddela-etal-2023-lens}.}
\label{table:multi_model}
\end{table}

\cam{We include additional results on our main experiments in in Appendix \ref{appdx:further_results}, notably that multi-prompt outperforms beam search and that the choice of the single prompt impacts the baseline performance.}

\subsection{What is the impact of the prompt bank?}
\label{sec:prompt_sampling}

\pg{Sampling Prompts Improves Candidate Quality.} 
Table \ref{table:prompt_selection} (top) reports results for multi-prompt across different prompt sampling methods for text simplification and translation. 
We perform a hypothesis test for the statistical significance of each variation of multi-prompt outperforming single prompt MBR using bootstrap sampling with 1000 iterations \citep{koehn-2004-statistical}.
Note that, code generation results are omitted as a unique set of prompts is generated for each HumanEval example.
% , rather than the same prompts used across the entire dataset.
We find sampling prompts by usage and truncating the top-$p$ prompts improves multi-prompt over a random selection baseline, with top-$p$ prompt sampling performing the best on both tasks. 

\vspace{2pt}
\pg{A Higher Quality Prompt Bank Improves Multi-prompt.}
Table \ref{table:prompt_selection} (bottom) reports results for different prompt subset selection methods, which use heuristics to select a smaller set of prompts for multi-prompt to maximize performance. 
The best selection method for each task had a significant impact on performance when compared to a single prompt MBR (+2.9 pass@1, +4.9 \textsc{Lens} and +0.9 \textsc{Comet}).
For text simplification, decoding with the 10 highest performing prompts is further improved by selecting prompts from a $k$-NN clustering of prompt embeddings, which enforces a dis-similarity between prompts. 
However, translation and code generation benefit from using the farthest similarity, or semantically distant prompts. 
These results highlight multi-prompt's sensitivity to the prompt construction, and shows that enforcing both diversity via multi-prompt and performance via prompt selection improves candidate generation. A direct comparison between prompt sampling and selection using the same candidate set size is included in Table \ref{table:prompt_selection_detailed} in Appendix \ref{appdx:prompt_selection_detailed}.

\begin{figure}[t!]
\centering
\includegraphics[width=0.47\textwidth]{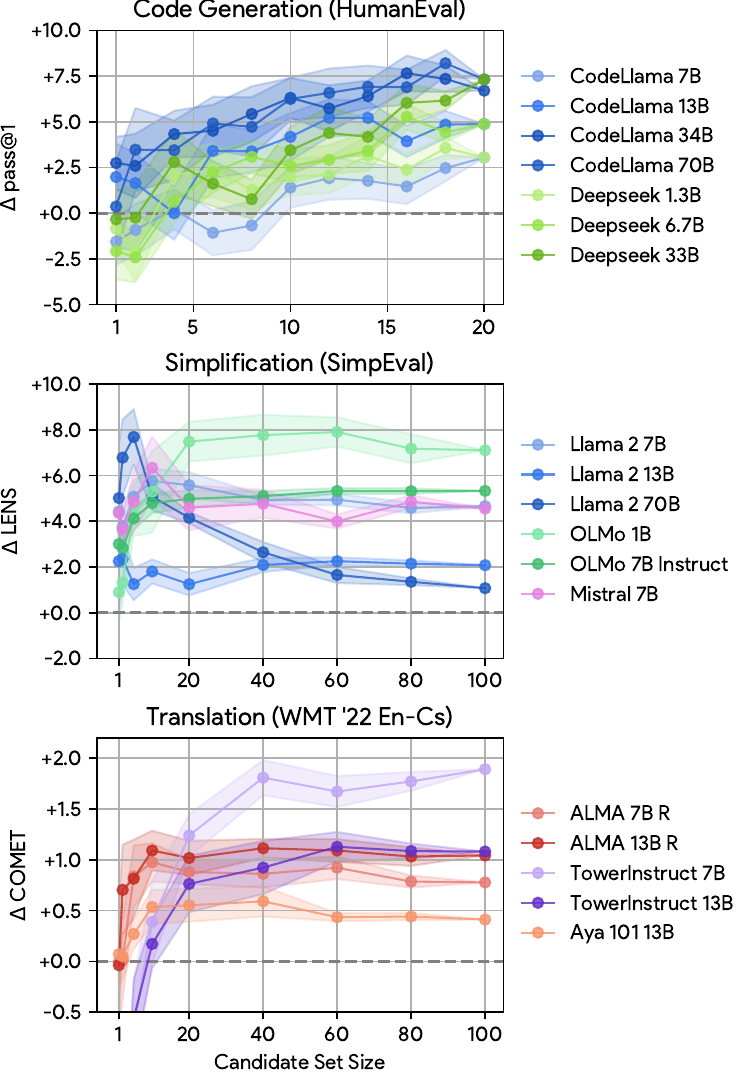}
\setlength{\belowcaptionskip}{-10pt}
\captionof{figure}{$\Delta$ metric improvement from single prompt to multi-prompt across model sizes and architectures, reported with a 95\% CI bootstrapped over 20 iterations. For absolute performance, see Figure \ref{fig:multi_prompt_detailed}.}

\label{fig:multi_model}
\end{figure}

% \subsection{Scaling Multi-Prompt}
\subsection{Does multi-prompt MBR improve quality across model architectures and sizes?}
\label{sec:multi_model}

\vspace{2pt}
\pg{Multi-prompt Improves MBR Across Models.} 
% To argue multi-prompt improves generation across instruction fine-tuned models and at scale, we experiment with widely used LLMs.
Figure \ref{fig:multi_model} reports improvement of multi-prompt over single prompt across widely used LLMs as a $\Delta$ change in score, with per-model results in Appendix \ref{appdx:multi_model_results}. 
In all cases, multi-prompt outperforms single prompt using a sufficiently large candidate set size, showing an increasing or constant metric improvement. In fact, smaller models surpass their larger counterparts' single output decoding at large enough candidate set sizes (Fig. \ref{fig:multi_prompt_detailed}). For instance, CodeLlama 13B outperforms its 70B variant using multi-prompt with 18 candidates ($48.26\!>\!47.99$ pass@1) and TowerInstruct 7B outperforms 13B with 5 candidates ($81.73\!>\!80.14$ \textsc{Comet}).

% On code generation, models saw increasing returns using multi-prompt as candidate set size increased, and translation models converged to a consistent gain across all models.
% For text simplification, we find inconsistent or closing gap in improvement for larger scale models.
% However, in all cases, multi-prompt improved single prompt after an adequate candidate set size.

\vspace{2pt}
\pg{LLMs with Multi-prompt Outperform Fine-tuned Models.} 
Whether general-purpose, instruction fine-tuned LLMs outperform models trained on a specific generation task is still an active question \citep{qin2023chatgpt}, so we compare state-of-the-art results from each task dataset using single prompt MBR to instruction fine-tuned LLMs using multi-prompt MBR with top-$p$ prompt sampling. 
In Table \ref{table:multi_model}, we report previous SOTA results for each task: an 11B T5-based text simplification model 
with control tokens for simplification operations 
\citep{sheang-saggion-2021-controllable}, the \textsc{En-Cs} results for the WMT '22 winning submission \citep{kocmi-etal-2022-findings} and StarCoder 15B, a code infilling and generation LLM \citep{li2023starcoder}, not explicitly trained to follow natural language instructions. 
LLMs surpass fine-tuned model performance when using multi-prompt, for instance Llama 2 13B shows +5.8 \textsc{Lens} over fine-tuned T5 11B. 

\begin{figure*}[t!]
\centering
\includegraphics[width=0.98\textwidth]{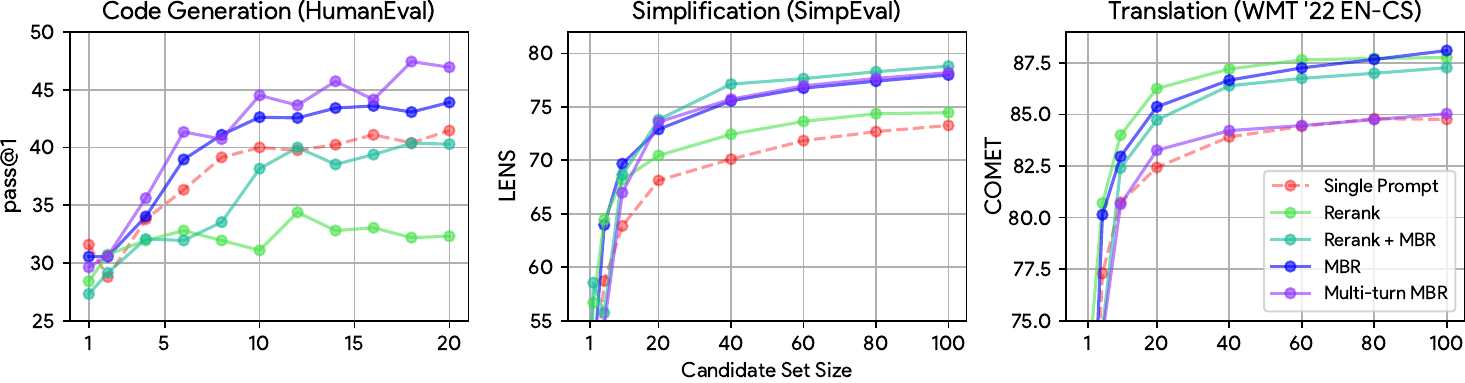}
\setlength{\abovecaptionskip}{5pt}
\setlength{\belowcaptionskip}{-10pt}
\captionof{figure}{Alternative MBR formulations for multi-prompt across candidate set sizes for code generation, text simplification and translation. Efficient MBR methods show inconsistent results, dependent on task and metric.}
\label{fig:rerank}
\end{figure*}
\begin{table}[t!]
\centering
\fontsize{8}{10}\selectfont
\begin{tikzpicture}
\node (table) {
\begin{tabular}{L{58pt}C{20pt}C{20pt}C{20pt}C{20pt}C{20pt}C{20pt}}

\\[-2ex]
\multicolumn{7}{c}{\textit{Text Simplification (LLaMA 7B Chat)}} \\
\hdashrule

& 
{\hspace{-0.8cm}\rotatebox[origin=r]{-30}{{\textsc{Sari}}}} &
{\hspace{-1.3cm}\rotatebox[origin=r]{-30}{\textsc{BertScore}}} & 
{\hspace{-0.7cm}\rotatebox[origin=r]{-30}{\textsc{Lens}}} & 
{\hspace{-1.4cm}\rotatebox[origin=r]{-30}{\textsc{Lens-Salsa$^{\text{rf}}$}}} & 
{\hspace{-0.4cm}\rotatebox[origin=r]{-30}{{\textsc{Sle$^{\text{rf}}$}}}} \tabularnewline
\midrule

\textsc{Sari} & \cellcolor{green!25} +1.08$^*$ & \cellcolor{green!25} +1.06$^*$ & \cellcolor{green!25} +7.24$^*$ & \cellcolor{green!25} +4.33$^*$ & \cellcolor{green!25} +0.38$^*$ \\
\textsc{BertScore} & \cellcolor{green!25} +1.44$^*$ & \cellcolor{green!25} +1.09$^*$ & \cellcolor{green!25} +6.18$^*$ & \cellcolor{green!25} +3.11$^*$ & \cellcolor{green!25} +0.45$^*$ \\
\textsc{Lens} & \cellcolor{red!25} -0.67 & \cellcolor{red!25} -0.05 & \cellcolor{green!25} +5.78$^*$ & \cellcolor{green!25} +4.69$^*$ & \cellcolor{green!25} +0.82$^*$ \\
\textsc{Lens\nobreakdash-Salsa$^{\text{rf}}$} & \cellcolor{red!25} -0.83 & \cellcolor{green!25} +0.35$^*$ & \cellcolor{green!25} +8.10$^*$ & \cellcolor{green!25} +4.65$^*$ & \cellcolor{green!25} +0.97$^*$ \\
\textsc{Sle$^{\text{rf}}$} & \cellcolor{red!25} -5.25 & \cellcolor{red!25} -4.71 & \cellcolor{green!25} +2.39$^*$ & \cellcolor{red!25} -4.51 & \cellcolor{green!25} +1.05$^*$ \\

\bottomrule

\rule{0pt}{2.5ex}
& \multicolumn{6}{l}{\textit{Translation (ALMA 7B)}} \\
\hdashrule

& 
% {\hspace{-1.0cm}\rotatebox[origin=r]{-30}{\textsc{Bleu}}} &
{\hspace{-1.4cm}\rotatebox[origin=r]{-30}{\textsc{BertScore}}} &
{\hspace{-1.2cm}\rotatebox[origin=r]{-30}{\textsc{Comet-22}}} &
{\hspace{-1.4cm}\rotatebox[origin=r]{-30}{\textsc{CometKiwi$^{\text{rf}}$}}} & {\hspace{-0.8cm}\rotatebox[origin=r]{-30}{\textsc{xComet}}} &
{\hspace{-0.8cm}\rotatebox[origin=r]{-30}{\textsc{MetricX}}} &
{\hspace{-1.4cm}\rotatebox[origin=r]{-30}{\textsc{MetricX-QE$^{\text{rf}}$}}} \tabularnewline
\midrule

\textsc{Bleu} & \cellcolor{green!25} +0.34$^*$ & \cellcolor{green!25} +0.47$^*$ & \cellcolor{green!25} +0.67$^*$ & \cellcolor{red!25} -0.14 & \cellcolor{green!25} +0.04 & \cellcolor{green!25} +0.11$^*$ \\
\textsc{BertScore} & \cellcolor{green!25} +0.51$^*$ & \cellcolor{green!25} +1.59$^*$ & \cellcolor{green!25} +1.68$^*$ & \cellcolor{green!25} +2.48$^*$ & \cellcolor{green!25} +0.22$^*$ & \cellcolor{green!25} +0.29$^*$ \\
\textsc{Comet\nobreakdash-22} & \cellcolor{green!25} +0.71$^*$ & \cellcolor{green!25} +0.89$^*$ & \cellcolor{green!25} +1.72$^*$ & \cellcolor{green!25} +3.29$^*$ & \cellcolor{green!25} +0.13$^*$ & \cellcolor{green!25} +0.18$^*$ \\
\textsc{CometKiwi$^{\text{rf}}$} & \cellcolor{green!25} +0.80$^*$ & \cellcolor{green!25} +1.03$^*$ & \cellcolor{green!25} +1.06$^*$ & \cellcolor{green!25} +2.87$^*$ & \cellcolor{green!25} +0.07$^*$ & \cellcolor{green!25} +0.08$^*$ \\
\textsc{xComet} & \cellcolor{green!25} +0.14 & \cellcolor{green!25} +0.85$^*$ & \cellcolor{green!25} +0.84$^*$ & \cellcolor{green!25} +3.34$^*$ & \cellcolor{green!25} +0.09$^*$ & \cellcolor{green!25} +0.04$^*$ \\
\textsc{MetricX} & \cellcolor{green!25} +0.36$^*$ & \cellcolor{green!25} +0.81$^*$ & \cellcolor{green!25} +0.36 & \cellcolor{green!25} +3.93$^*$ & \cellcolor{green!25} +0.07$^*$ & \cellcolor{red!25} -0.04 \\
\textsc{MetricX\nobreakdash-QE$^{\text{rf}}$} & \cellcolor{green!25} +0.60$^*$ & \cellcolor{green!25} +1.68$^*$ & \cellcolor{green!25} +2.11$^*$ & \cellcolor{green!25} +5.31$^*$ & \cellcolor{green!25} +0.08$^*$ & \cellcolor{green!25} +0.03$^*$ \\

\bottomrule
\end{tabular}
};
\coordinate (nw) at (table.north west);
\coordinate (ne) at (table.north east);
\coordinate (sw) at (table.south west);
\draw[thick, -latex] ($(nw) + (0, 0.25em)$) -- ($(ne) + (0, 0.25em)$)
node[midway, fill=white] {Evaluation Metric};
\draw[thick, -latex] ($(nw) + (-0.5em, 0)$) -- ($(sw) + (-0.5em, 0)$)
node[midway, fill=white, rotate=90] {MBR Utility Metric};
\end{tikzpicture}
\setlength{\belowcaptionskip}{-10pt}
\setlength{\abovecaptionskip}{-4pt}
\caption{$\Delta$ metric improvement from single prompt to multi-prompt across metrics. \textsc{rf $=$} Reference-free reranker. * $=$ Statistically significant improvement with $p<0.05$. For absolute performance, see Table \ref{table:cross_metric_detailed}.}
\label{table:cross_metric}
\end{table}

\vspace{2pt}
\pg{Candidate Set Overlap May Explain the Performance Similarity for Large Models.}
\cam{
Finally, in Table \ref{table:multi_model}, we observe that stronger systems, such as GPT-4 on translation, show smaller differences between single and multi-prompt. 
One explanation may be due to stronger models generating similar candidate sets between both methods.
To understand this behavior, we measure the similarity between the candidate set generated by multi-prompt and single prompt, where a higher similarity candidate set may indicate a smaller improvement from multi-prompt. 
We report the `Candidate \textsc{Bleu} (\texttt{target} on \texttt{references})' score, which measures of the $n$-gram overlap of a set of target sequences over the bank of references.
In our results, we find that stronger models produce single prompt candidate sets which contain more multi-prompt $n$-grams (as shown in `SP on MP'), and that candidate sets show a higher $n$-gram coverage as models improve.
This increasing similarity between the candidates may explain the decreasing performance improvement for multi-prompt.
}

\subsection{Does multi-prompt MBR over-fit to the utility metric?}
\label{sec:metric_eval}
An inherent challenge of evaluating MBR is that the utility metric used to select candidates is typically also used for the final evaluation, in such cases it is difficult to attribute the metric improvement to higher quality generation \citep{bertsch-etal-2023-mbr}. Given growing attention to metric development, we leverage various trained metrics to test whether multi-prompt using one utility metric improves performance cross all other utility metrics. We experiment with traditional overlap-based metrics, (\textsc{Bleu}, \textsc{Sari}), embedding similarity (\textsc{BertScore}), small ($\sim$100M parameter) trained metrics with references (\textsc{Lens}, \textsc{Comet-22}) and without references (\textsc{CometKiwi}, \textsc{Lens-Salsa}, \textsc{Sle}), and large (3B+ parameter) trained metrics (\textsc{xComet}, \textsc{MetricX}, \textsc{MetricX-QE}). These metrics represent diverse text evaluation approaches and encompass the full state of evaluation in both tasks. We include a full description of metric architectures in Appendix \ref{appdx:metrics}.
\vspace{2pt}

\pg{Multi-prompt MBR Improves Across Metrics.}
Table \ref{table:cross_metric} reports results for cross-metric evaluation, with the diagonal reflecting the traditional MBR evaluation setup (i.e., calculate MBR and evaluate using the same metric) and other cells indicate generalization from one metric to all others. 
Multi-prompt improves performance on most evaluation setups, with a few notable exceptions such as disagreement between trained and overlap-based metrics for simplification and \textsc{Comet}-based metrics for translation.
For simplification, trained metrics' failure when evaluated by \textsc{Sari} and \textsc{BertScore} may be a byproduct of the test set size, as these metrics typically require a substantial number of references for stable evaluation \citep{alva-manchego-etal-2020-asset}, more than what are provided in \textsc{SimpEval}.
Interestingly, the magnitude of performance improvement is highly variable to the specific utility metric, with no clear relationship between the metric architecture and improvement of multi-prompt, but typically a lower baseline performance indicates multi-prompt performs better (Table \ref{table:cross_metric_detailed} in Appendix for more details).

\subsection{How does the metric type impact multi-prompt MBR?}
\label{sec:rerank}
As discussed by \citet{fernandes-etal-2022-quality}, the MBR operation requires each candidate evaluate against every other candidate (i.e., $\mathcal{O}(n^2)$ comparisons), this becomes inefficient in practice for a large $n$, especially when using a trained utility metric. Therefore, we explore multi-prompt MBR alternatives using reference-free utility metrics:

\begin{itemize}[noitemsep,topsep=0.5pt,leftmargin=*]
\setlength\itemsep{0.5pt}
\item \textbf{Reranker ($\mathcal{O}(n)$).} Re-ranking directly estimates the quality of each candidate using a reference-free metric: $\hat{y}_{\text{MBR}}\!=\!\argmax_{y\in \mathcal{H}} \left[ \text{U} (y)\right]$. We use the trained \textsc{Lens-Salsa} for simplification \citep{heineman-etal-2023-dancing} and \textsc{Comet-MQM} \citep{rei-etal-2021-references} for translation. For code generation, we use Code Reviewer \citep{shi-etal-2022-natural}, which calculates agreement between the per-token probability of the generation given the docstring and the original docstring given the generation. Reference-free re-ranking only requires $n$ metric calculations to directly estimate quality.
\item \textbf{Reranker + MBR ($\mathcal{O}(n+m^2)$).} We use a two-stage selection where we first rerank all $n$ candidates and select the top $m$ to use for MBR, where the cheap re-ranker can distill the candidate set and the expensive MBR metric performs the final selection, where $m\ll n$.
\item \textbf{Multi-turn MBR ($\mathcal{O}(n^2+m^2)$).} Similar to the previous approach, we perform MBR and then re-compute MBR using the top $m$ candidates.
\end{itemize}
\vspace{1pt}

\pg{Results.} 
We report results across candidate selection methods in Figure \ref{fig:rerank}, finding the multi-prompt achieves performance improvement across reference-based and reference-free metrics, yet the relative performance of methods varies between tasks. 
With text simplification, the methods first narrowing the candidate set (`Rerank + MBR') and iteratively performing MBR (`Multi-turn MBR') either match or out-perform vanilla MBR. 
We speculate the first pass may prune the lowest quality generations such that the second pass only considers a distilled candidate set, which better informs the MBR calculation.
For translation, the more efficient re-ranker outperforms vanilla MBR, which follows recent work finding trained reference-based and reference-free MT metrics are approaching a similar quality \citep{freitag-etal-2023-results}.
For code generation, the re-ranker under-performs MBR, which may be reflective of the performance of Code Reviewer compared to \textsc{MBR-Exec}, as the latter has access to multiple test cases.

\section{Related Work}
\label{sec:related_work}
\pg{Output Selection.}
Ensembling outputs across a generation set has become a widely used technique for improving LLM performance in classification tasks, such as using a majority vote over reasoning chains \citep{wang2023selfconsistency}, or merging outputs from multiple models \citep{kobayashi-2018-frustratingly, martinez-lorenzo-etal-2023-amrs}. 
This work applies the same underling concept to text generation by leveraging trained automatic evaluation metrics.
To our knowledge, it is the first to propose a multi-prompt decoding scheme for text generation.
\vspace{2pt}

\pg{MBR Decoding.}
MBR decoding has been previously used to improve generation quality for machine translation \citep{kumar-byrne-2004-minimum, eikema-aziz-2020-map, muller-sennrich-2021-understanding} text simplification \citep{maddela-etal-2023-lens}, summarization and style transfer \citep{suzgun-etal-2023-follow}. 
\citet{bertsch-etal-2023-mbr} highlight the growing popularity of MBR as a simple technique in machine translation and reporting shared tasks results.
While our work is the first to propose generating the MBR hypothesis space using a prompt bank, \citet{farinhas2023empirical} perform preliminary experiments with paraphrases of a single sentence prompt, but found no difference in performance. 
Recent work argues sampling strategies like nucleus \citep{eikema-aziz-2022-sampling} or epsilon \citep{freitag-etal-2023-epsilon} offer slightly better performance over beam search for MBR, with this work extending their findings by attributing candidate set quality to sampling diversity. 
\vspace{2pt}

\pg{Prompt Selection.}
Current work on prompting for text generation has instead focused on optimization, such as in-context example selection \citep{min-etal-2022-rethinking}, example ordering \citep{lu-etal-2022-fantastically} and prompt selection \citep{gonen-etal-2023-demystifying}.
Notably, \citet{agrawal-etal-2023-context} show selecting in-context examples for MT by maximizing $n$-gram overlap between the source and  examples improves few-shot performance. 
\citet{zhou2023large} experiment with LLMs as prompt generators, and \citet{yang2023large} show using LLMs to iteratively rewrite prompts on a development set can distill a single, high-performant prompt. 
Our work builds on LLM-written prompts and basic heuristics for distilling the prompt bank to further improve multi-prompt.
\vspace{2pt}

\section{Conclusion}
In this work, we propose multi-prompt, a generalized case of MBR for conditional text generation.
Multi-prompt successfully ensembles outputs of instruction fine-tuned language models across prompt constructions and in-context examples. 
We highlight the importance of prompt selection and sampling when constructing the prompt bank with top-$p$ prompt sampling and further verify our results across tasks, models and utility metrics.

\section*{Limitations}
We limit our study of the prompt bank to a basic set of seed prompts and GPT-written paraphrases.
Notably, we do not study the impact of prompt formats (e.g., \texttt{passage:\{\}\textbackslash n answer\{\}} vs. \texttt{Passage::\{\} Answer::\{\}}, \citealp{sclar2023quantifying}), in-context example ordering \citep{lu-etal-2022-fantastically} or example selection \citep{agrawal-etal-2023-context} on multi-prompt performance, although multi-prompt may extend to such methods. 
We leave the question of exhaustively constructing a prompt bank to future work.

An inherent limitation of MBR is the increase in inference time, where we generate up to 500 samples in our experiments, and use a neural utility metric with either linear or quadratic comparisons between candidates. 
\cam{To illustrate this, the wall clock time for the main experiment setup (Figure \ref{fig:multi_prompt_small}) using standard decoding on a single A40 GPU is 4.73, 2.10, 2.21 seconds per input sentence and for multi-prompt with 100 candidates is 38.76, 183.81, 124.70 seconds per input sentence for code generation, simplification and translation respectively.}

In practice, the generation time was significantly lowered by decoding in parallel and the use of efficient-memory attention techniques such as paged and flash attention used in the vLLM library \citep{kwon2023efficient}. 
The computational bottleneck for large candidate set sizes was instead evaluating the utility metrics across all pairs of generated candidates.
To lower the number of metric comparisons, promising results have been demonstrated by pruning low-scoring candidates during the MBR process \citep{cheng-vlachos-2023-faster}, aggregating embedding representations of candidates \citep{vamvas2024linear} or selecting a subset of references for each candidate using heuristics on reference embeddings \citep{deguchi2024centroid}. 
Similarly, we show in \S\ref{sec:rerank} efficient alternatives to MBR such as using reference-free metrics largely preserve the benefits from multi-prompt.

Along with MBR, many widely used methods improving LLM abilities trade increased compute at inference time for higher performance, such as using chain-of-thought to decode a reasoning chain for a single answer or using self-consistency to selects an answer among multiple reasoning chains \citep{wei2022chain, wang2023selfconsistency}.

\section*{Acknowledgments}
The authors would like to thank Alan Ritter and Y-lan Boureau for discussions and Duong Le for his feedback on a draft manuscript. This research is supported in part by the NSF awards IIS-2144493 and IIS-2112633, NIH award R01LM014600, ODNI and IARPA via the HIATUS program (contract 2022-22072200004). The views and conclusions contained herein are those of the authors and should not be interpreted as necessarily representing the official policies, either expressed or implied, of NSF, NIH, ODNI, IARPA, or the U.S. Government. The U.S. Government is authorized to reproduce and distribute reprints for governmental purposes notwithstanding any copyright annotation therein.

% \bibliography{cite/anthology_p1, cite/anthology_p2, cite/custom}

\clearpage

\appendix

\section{Prompt Bank Construction}
\label{appdx:prompts}
Table \ref{table:human_written_prompts} contains the human-written prompts for text simplification. These human-written prompts are provided as examples to GPT-4 when automatically generating prompts for large-scale experiments in \S\ref{sec:results}. For code generation, we extract the docstring in the original \textsc{HumanEval} examples as the human-written prompt, and provide it as an example prompt to GPT-4. For machine translation, our few-shot examples were sampled randomly from the WMT \texttt{newstest19} test corpus \citep{barrault-etal-2019-findings}.

\begin{table}[t!]
\centering
\tiny
\begin{tabular}{p{0.95\linewidth}}
\toprule
\textbf{Human-Written Text Simplification Prompt} \\
\midrule
I am writing a sentence, please take a look at this sentence and write a simpler version such that a non-english speaker or an individual with disabilities could better understand the sentence. \\
\midrule
Rewrite the following complex sentence in order to make it easier to understand by non-native speakers of English. You can do so by replacing complex words with simpler synonyms (i.e. paraphrasing), deleting unimportant information (i.e. compression), and/or splitting a long complex sentence into several simpler ones. The final simplified sentence needs to be grammatical, fluent, and retain the main ideas of its original counterpart without altering its meaning. \\
\midrule
You are an artificial intelligence designed to simplify human written text. The text you are given will contain complex ideas, phrases or concepts and your job is to rewrite that text in a simple and easy to understand way. Your simplification should be completely fluent and retain the ideas of the simplification. \\
\midrule
I would like you to simplify the following sentence such that the text is as concise and easy to read as possible. \\
\midrule
You are to act as a text simplification bot. As a text simplification bot, you will simplify the following sentence such that it is syntactically easier to read and semantically easier to understand. Please do not make the text more complex, longer or difficult for a reader. \\
\midrule
Make this sentence more approachable for a non-english speaker or an individual with a disability. \\
\midrule
Rewrite the following sentence in simpler terms to help non-native English speakers and people with disabilities understand it better. \\
\midrule
This is a sentence from Wikipedia, rewrite it such that it could appear on Simple English Wikipedia \\
\midrule
You are an AI assistant that writes text simplification. Text simplification can be defined as any process that reduces the syntactic or lexical complexity of a text while attempting to preserve its meaning and information content. The aim of text simplification is to make text easier to comprehend for a human user, or process by a program. Please simplify the following sentence. \\
\midrule
The following sentence has a high CEFR rating. Can you please rewrite it such that it will have a lower CEFR classification? \\
\bottomrule
\end{tabular}
\setlength{\belowcaptionskip}{-10pt}
\caption{Text simplification prompts used for the decoding experiment in Figure \ref{fig:candidate-probabilities} and used as examples to write GPT-4 prompts for experiments in \S\ref{sec:results}.}
\label{table:human_written_prompts}
\end{table}

\section{Detailed System Descriptions}
\label{appdx:system_details}
In this section, we include a full description of the generation models and utility metrics used in experiments throughout \S\ref{sec:multi_model} and \S\ref{sec:metric_eval}. All experiments were inference-based and were run on up to 4xNVIDIA A40 GPUs, depending on the requirements of the specific model or utility metric. The use of models, metrics and datasets in this project follows their respective licenses and intended use.

\begin{table}[t!]
\centering
\tiny
\begin{tabular}{p{0.95\linewidth}}
\toprule
\textbf{Prompt-Generation Instruction} \\
\midrule

Please write a variation of the following instruction for a coding task. You may be creative in proposing potential solutions, or explaining the nature of the task. Please do not write any examples. \\[3pt]

Example: \texttt{\{example\_prompt\}} \\[3pt]

Prompt: \\
\midrule

Create a prompt for a language model to simplify a sentence, this prompt will explain the text simplification task and instructions for how to perform the task. The prompt should be diverse, include a description of simplification and clearly state what is expected of the language model. \\[3pt]

Example: \texttt{\{example\_prompt\_1\}} \\[3pt]

Example: \texttt{\{example\_prompt\_2\}} \\[3pt]

Prompt: \\

\bottomrule
\end{tabular}
\setlength{\belowcaptionskip}{-10pt}
\caption{Instruction templates provided to GPT-4 when generating task instructions for code generation (top) and text simplification (bottom).}
\label{table:prompt_generation}
\end{table}

\subsection{Utility Metrics}
\label{appdx:metrics}

\subsubsection{Code Generation}

\vspace{2pt}
\pg{\textsc{MBR-Exec} \citep{shi-etal-2022-natural}} executes candidate generations on a series of test cases, and selects the candidate with the highest agreement on its output with all other candidates. While the authors do not evaluate on \textsc{HumanEval}, we replicate the setup in \citet{zhang2023coder} by using the test cases in the docstring to calculate the agreement. We use a soft loss over all test cases, as many \textsc{HumanEval} docstring examples are trivial or edge cases. If two candidates have the same MBR score, we break ties using the candidate with higher probability under the language model.

\pg{Code Reviewer \citep{zhang2023coder}} attempts to find a consensus between the likelihood of the generated program $p(y|x)$ and the original docstring using a minified version of the generation $p(x|y)$. We use their implementation for rejecting degenerate samples, minifying code and calculating the reviewer score. We use the same models for generation and re-ranking.

\subsubsection{Simplification}

\pg{\textsc{Sari} \citep{xu-etal-2016-optimizing}} is an $n$-gram overlap based metric that compares edits on inputs, outputs and a bank of references.

\vspace{2pt}
\pg{\textsc{BERTScore} \citep{zhang2019bertscore}} calculates a word-level cosine similarity of BERT embeddings. \citet{alva2021suitability} find \textsc{BERTScore} is an adequate measure of quality generation, but that it does not correlate with simplicity.

\vspace{2pt}
\pg{\textsc{Lens} \citep{maddela-etal-2023-lens}} is a RoBERTa-based metric trained using human ratings of text simplification model outputs. The authors train on an adaptive loss to allow a high score for generations that are close to \textit{any} references, encouraging the metric to consider different simplification types.

\vspace{2pt}
\pg{\textsc{Lens-Salsa} \citep{heineman-etal-2023-dancing}} extends the \textsc{Lens} architecture by fine-tuning on a dual sentence- and word-level quality objective. The authors show \textsc{Lens-Salsa} is more sensitive to specific edit operations, while not requiring any reference simplifications.

\vspace{2pt}
\pg{\textsc{Sle} \citep{cripwell-etal-2023-simplicity}} is a RoBERTa-based metric trained to estimate the simplicity of text, with the simplicity score defined as the difference in simplicity between the complex and simplified sentences. \textsc{Sle} was trained on 0-4 readability scores of news articles in the Newsela corpus \citep{xu-etal-2015-problems}, with an additional label softening for individual sentences in each article.

\subsubsection{Translation}

\pg{\textsc{Bleu} \citep{papineni-etal-2002-bleu}} is an $n$-gram overlap based metric comparing a translation to a bank of references. BLEU remains a widely-used standard for automatic evaluation, despite lower correlation to human judgement compared to learned metrics \citep{freitag-etal-2022-results}. We use the ScareBLEU implementation \citep{post-2018-call}.

\vspace{2pt}
\pg{\textsc{Comet} \citep{rei-etal-2020-comet}} is a widely used RoBERTa-based metric, trained on direct assessments of simplification quality. For reference-free evaluation, we use the CometKiwi-XXL variant \citep{rei-etal-2022-cometkiwi, rei-etal-2023-scaling}, trained to predict sentence- and word-level scores simultaneously.

\vspace{2pt}
\pg{\textsc{xComet} \citep{guerreiro2023xcomet}} is a fine-tuned XLM-R model \citep{goyal-etal-2021-larger} based on the CometKiwi architecture, but scaling the model size and training data, including with synthetic data created by randomly swapping $n$-grams or entire sentences with unrelated translations. We use the 11B \textsc{xComet-xxl} in our experiments.

\vspace{2pt}
\pg{\textsc{MetricX} \citep{juraska-etal-2023-metricx}} is a recent fine-tuned 11B mT5-XXL \citep{xue-etal-2021-mt5} trained on DA data from 2015-20, MQM data from 2020-21 \citep{freitag-etal-2021-experts} and synthetic data based on the MQM and DEMETR \citep{karpinska-etal-2022-demetr} taxonomies of translation errors. Notably, the MetricX architecture encodes both candidates and references together, while COMET encodes both separately and combines the outputs to calculate the final score. We also use the reference-free variant \textsc{MetricX-QE}. The WMT '22 test data used in this work is not included in the training data of any translation metrics we considered. 

\subsection{Model Architectures}
\label{appdx:models}

\subsubsection{Code Generation}

\pg{StarCoder 2 \citep{li2023starcoder}} is trained from-scratch on 4T tokens from 600+ programming languages. Although the model is not instruction fine-tuned, we see a slight performance improvement with multi-prompt, likely because comments and code descriptions are included in its pre-training.

\vspace{2pt}
\pg{CodeLLaMA \citep{roziere2023code}} is a fine-tuned Llama 2 model on 500B-1T tokens of code-related datasets, including Python, substantially outperforming the base Llama 2 model on HumanEval.

\subsubsection{Simplification}

% \pg{Llama 2 \citep{touvron2023llama}}
\pg{Instruction Fine-tuned Models.}
We experiment with widely used instruction fine-tuned LLMs, aiming for a broad coverage of current models: Llama 2 Chat \citep{touvron2023llama}, Gemma \citep{team2024gemma} and Mistral \citep{jiang2023mistral}. 

\vspace{2pt}
\pg{Fine-tuned Control T5 \citep{sheang-saggion-2021-controllable}} is a T5-based text simplification model fine-tuned on the Wiki-Auto \citep{jiang-etal-2020-neural} dataset of aligned English-Simple English Wikipedia articles. We use their same control token setup: \texttt{<NC\_0.95> <LS\_0.75> <DR\_0.75> <WR\_0.75>}.

\begin{figure*}[t!]
\centering
\includegraphics[width=0.98\textwidth]{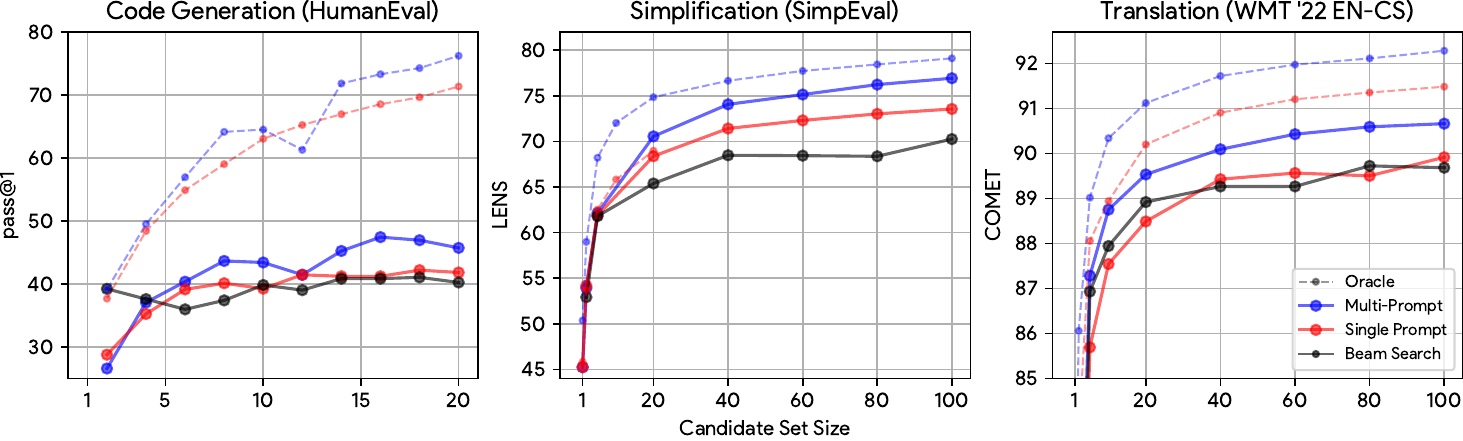}
\setlength{\abovecaptionskip}{8pt}
\setlength{\belowcaptionskip}{-5pt}
\captionof{figure}{\textcolor{blue}{Multi-prompt}, \textcolor{red}{single prompt} and beam search MBR decoding performance across candidate set sizes for code generation, text simplification and translation. Results are an average over 5 repetitions.}
\label{fig:multi_prompt}
\end{figure*}

\subsubsection{Translation}

\pg{\textsc{Alma-r} \citep{xu2024contrastive}} is a class of translation LLMs. The base \textsc{Alma} \citep{xu2023paradigm} is a fine-tuned LLaMA model trained on monolingual text in each target language and further trained using parallel data. \textsc{Alma-r} \citep{xu2024contrastive} is an extension trained on a contrastive preference loss on ratings of translation quality.

\vspace{2pt}
\pg{TowerInstruct \citep{alves2024tower}} is a fine-tuned Llama 2 model on multi-lingual instructions, aiming to incorporate tasks beyond translation, such as paraphrasing, post editing and grammar error correction.

\vspace{2pt}
\pg{Aya 101 \citep{ustun2024aya}} is an mT5-based model fine-tuned on multi-lingual data in 101 languages. While mT5 is an instruction-following model, Aya is not fine-tuned on instruction data.

Additionally, we provide results from the WMT '22 winning submission, and the Microsoft Translate API, as reported in \citet{hendy2023good}.

\newpage

\section{Further Results}
\label{appdx:further_results}
\subsection{Beam Search \& Oracle Performance}

Following related work in MBR, we report upper-bound `oracle' results (similar to \citealp{shi-etal-2022-natural}) and a lower-bound beam search baseline (similar to \citealp{freitag-etal-2023-epsilon}) in comparison to our main results (Figure \ref{fig:multi_prompt_small}) in Figure \ref{fig:multi_prompt}.

\vspace{2pt}
\noindent\textbf{Beam Search.} The MBR candidate set historically has consisted of the top beam search candidates, but as language models have become better generators recent work has argued sampling leads to a better estimation of the hypothesis space \cite{freitag-etal-2023-epsilon}. For this reason, we exclusively use nucleus sampling in \S\ref{sec:results}, but we report beam search as a baseline in Figure \ref{fig:multi_prompt}, with a `candidate set size' of $n$ corresponding to the top $n$ beam candidates, or $n$ candidates with nucleus sampling for other results.

\vspace{2pt}
\noindent\textbf{Oracle.} As the final MBR performance can be impacted both by the quality of the candidate set and the choice of utility metric, we report an upper-bound performance by deliberately selecting the best candidate generations. Given a test set with gold-standard references $\mathcal{R}$, we define the oracle performance as the set of the highest scoring possible selection of candidates:
\begin{equation}
\text{Oracle}(\mathcal{R}^*)=\sum_{r\in \mathcal{R}^{*}}\max_{y\in \mathcal{H}}\left[U(y, r)\right]
\end{equation}
Since code generation is evaluated using pass@1, its oracle uses expected pass@k \citep{shi-etal-2022-natural}, which measures whether at least one candidate within the candidate set passes all unit tests $\mathcal{T}$:
\begin{equation}
\text{ExPass@}K = \\
\underset{|\mathcal{H}|=K}{\mathbb{E}} \left[ \max_{y \in \mathcal{H}} \min_{t \in \mathcal{T}} \mathds{1} [t(y)] \right]
\end{equation}

\vspace{2pt}
\noindent\textbf{Results.} 
As oracle performance measures candidate set quality independent of the utility metric, we find an increase in oracle performance coincides with an improvement when using multi-prompt, indicating that a utility metric can naturally select candidates when the candidate set is higher quality. 
This suggests improving utility metrics may be a promising direction to bridge the gap between candidate quality and candidate selection. 
Beam search was a particularly strong baseline for small candidate set sizes, particularly for code generation, but beam search is not as sensitive to improvement as the candidate set size increases.
Additionally, as code generation is evaluated using the binary pass@1 metric, rather than a scalar quality metric as used by translation and simplification, there is a large gap between MBR and oracle performance, also observed by \citet{shi-etal-2022-natural}.

\begin{figure*}[t!]
\centering
\includegraphics[width=0.86\textwidth]{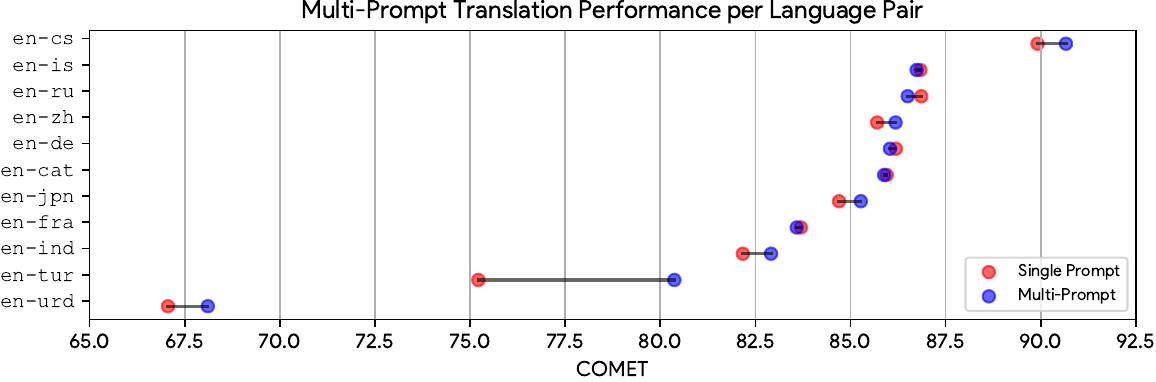}
\setlength{\abovecaptionskip}{5pt}
\setlength{\belowcaptionskip}{-5pt}
\captionof{figure}{\textcolor{blue}{Multi-prompt} and \textcolor{red}{single prompt} performance of ALMA 7B R across \texttt{En-XX} translation pairs. For low resource language pairs (e.g., Urdu, Turkish, Czech) we observe larger performance improvements compared to high resource pairs (e.g., French, German, Russian).}
\label{fig:language_pairs}
\end{figure*}

\subsection{\texttt{En-XX} Translation Results}
\label{appdx:translation_results}

For brevity, we limit our multi-prompt experiments to only the English-Czech language pair, but report results across the full ALMA test set, including WMT '22 test data and a subset of NTREX \citep{federmann-etal-2022-ntrex}, in Figure \ref{fig:language_pairs}, where we observe improvement with multi-prompt is dependent on the language pair.
Generally, high resource languages (such as French, German, Russian) do not have a substantial difference, which may be a result of the low prompt sensitivity for such pairs.
% This follows a similar findings in few-shot prompt selection for machine translation \citep{}. 

\subsection{Additional Multi-Prompt Results}
\label{appdx:multi_prompt_best}

\begin{figure}[t!]
\centering
\includegraphics[width=0.45\textwidth]{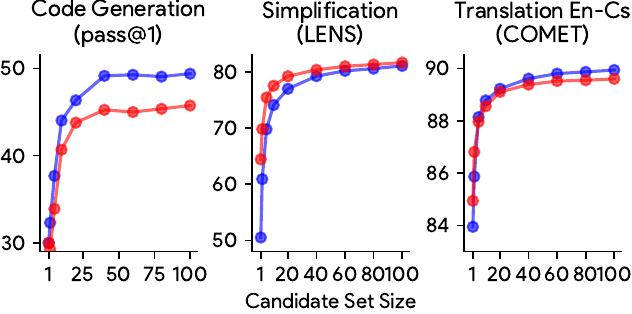}
\setlength{\belowcaptionskip}{-10pt}
\captionof{figure}{\textcolor{blue}{Multi-prompt} and \textcolor{red}{single prompt} MBR results from the setup in Figure \ref{fig:multi_prompt_small} with a different single prompt baseline. The single prompt was chosen as the highest usage $p(\rho)$ on the held-out dataset.}
\label{fig:multi_prompt_best}
\end{figure}

\cam{
In our main experiments, the single prompt setup uses a randomly selected prompt from the prompt bank. 
Instead, we experiment with using the prompt with the highest prompt usage $p(\rho)$ on the held-out 20\% of each dataset. 
In Figure \ref{fig:multi_prompt_best}, we report the performance of each method using the same setup as the main experiment (Figure \ref{fig:multi_prompt_small}) but using the alternative single prompt setup. 
For translation, we observe single-prompt and multi-prompt show a smaller performance difference. 
For text simplification, the highest usage prompt outperforms multi-prompt for small candidate sizes.
}

% \subsubsection{Analysis of Prompt Sampling}
% To provide intuition into how our prompt sampling impacts the set of prompts used to generate hypotheses, we run multi-prompt using 100 prompts, generating a single output from each prompt. Figure \ref{fig:prompt_distribution} reports the frequency that the generation from each prompt was selected as the MBR candidate among all prompts as a \% over the full dataset. A flat distribution indicates that each prompt equally produces the final MBR generation, but we find a few prompts receive disproportionately more frequent usage, with some prompts never producing the MBR candidate. Interestingly, both tasks have a very different distribution of usage, perhaps as translation is using few-shot examples, which may be less sensitive to performance as a natural language task instruction. 
% We show examples of the top prompts for simplification and translation in Table \ref{table:prompt_selection_examples}.

% \input{fig/appendix/prompt_distribution}}

\subsection{Additional Prompt Selection Results}
\label{appdx:prompt_selection_detailed}
To further compare prompt sampling and prompt selection with the same candidate set size, we replicate the same experiment as Table \ref{table:prompt_selection}, but modify prompt selection (bottom) to use 10 candidates for \textit{each} prompt, such that both sampling and selection use 100 candidates. We find similar results when comparing between prompt selection methods, where at least one selection method leads to a statistically significant improvement on each task. However, all prompt selection methods under-perform prompt sampling. This underscores the benefit of the increased diversity from generating using a full prompt bank with multi-prompt.

\setlength{\tabcolsep}{2pt}
\begin{table}[t!]
\centering
\fontsize{8}{9.5}\selectfont

\begin{tabular}{p{100pt}cC{1pt}cC{1pt}c}

\toprule
& pass@1 && \textsc{Lens} && \textsc{Comet} \\
\midrule

\textit{Single Prompt} ($|\mathcal{H}|\!=\!100$)  & 48.78 && 74.67 && 88.93 \\
\hdashrule
\multicolumn{6}{l}{\textit{Multi-Prompt + Prompt Sampling} ($|\mathcal{P}|\!=\!100$, $|\mathcal{H}|\!=\!100$)} \\

Random Selection        & -- && 74.91$^*$          && 89.98$^*$          \\
Prompt Sampling         & -- && 78.29$^*$          && 90.33$^*$          \\
Top-$p$ Prompt Random   & -- && 78.61$^*$          && 90.11$^*$          \\
Top-$p$ Prompt Sampling & -- && \textbf{79.08}$^*$ && \textbf{90.36}$^*$ \\

% \midrule

% \textit{Single Prompt} ($|\mathcal{H}|\!=\!10$)  & 41.55 && 51.64 && 87.54 \\
% \hdashrule
% \multicolumn{6}{l}{\textit{Multi-Prompt + Prompt Selection} ($|\mathcal{P}_{\text{best}}|\!=\!10$, $|\mathcal{H}|\!=\!10$)} \\

% Random Selection           & 39.63              && 60.00$^*$          && 87.81$^*$          \\
% $k$-NN Cluster Random      & 40.24              && 58.73$^*$          && 87.80$^*$          \\
% Farthest Similarity        & \textbf{44.51}$^*$ && 58.32$^*$          && \textbf{88.14}$^*$ \\
% Closest Similarity         & 37.80              && 61.53$^*$          && 87.73              \\
% Highest Performance        & --                 && 62.43$^*$          && 87.65              \\
% $k$-NN Cluster Performance & --                 && \textbf{66.12}$^*$ && 87.73              \\

\midrule

\textit{Single Prompt} ($|\mathcal{H}|\!=\!100$)  & 48.78 && 74.67 && 88.93 \\
\hdashrule
\multicolumn{6}{l}{\textit{Multi-Prompt + Prompt Selection} ($|\mathcal{P}_{\text{best}}|\!=\!10$, $|\mathcal{H}|\!=\!100$)} \\

Random Selection           & 47.40               && 70.95              && 89.90$^*$       \\
$k$-NN Cluster Random      & 45.73               && 72.04              && 90.14$^*$       \\
Farthest Similarity        & \textbf{49.17$^*$}  && 71.64              && 90.18$^*$       \\
Closest Similarity         & 45.73               && 72.17              && \textbf{90.87$^*$} \\
Highest Performance        & --                  && 72.56              && 90.27$^*$       \\
$k$-NN Cluster Performance & --                  && \textbf{75.88$^*$} && 90.43$^*$       \\

\bottomrule
\end{tabular}
\setlength{\belowcaptionskip}{-10pt}
\caption{Results for prompt sampling using 100 prompts (top) and subset selection with 100 candidates using 10 of 100 prompts (bottom). * $=$ Statistically significant improvement with $p\!<\!0.05$.}
\label{table:prompt_selection_detailed}
\end{table}

\subsection{Detailed Multi-Model Results}
\label{appdx:multi_model_results}

\begin{figure*}[t!]
\centering
\includegraphics[width=0.94\textwidth]{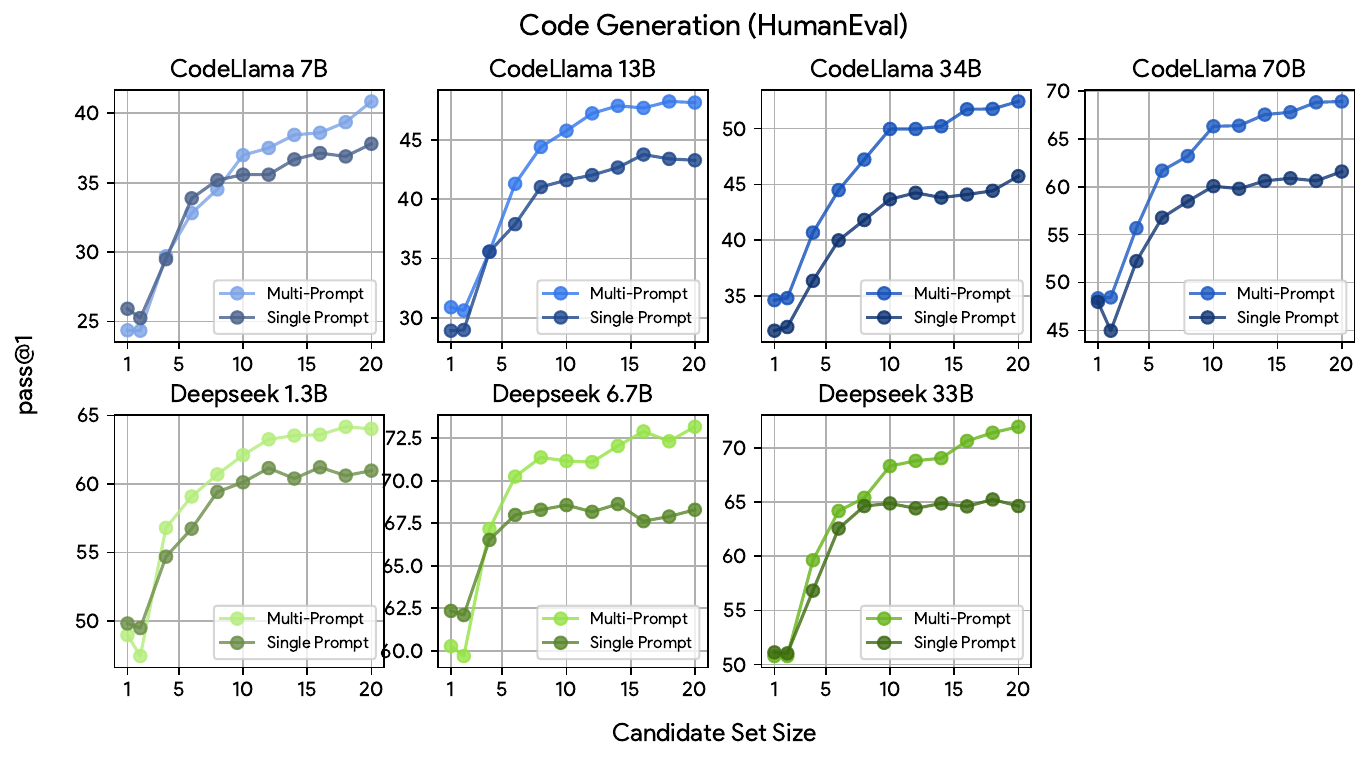}
\includegraphics[width=0.68\textwidth]{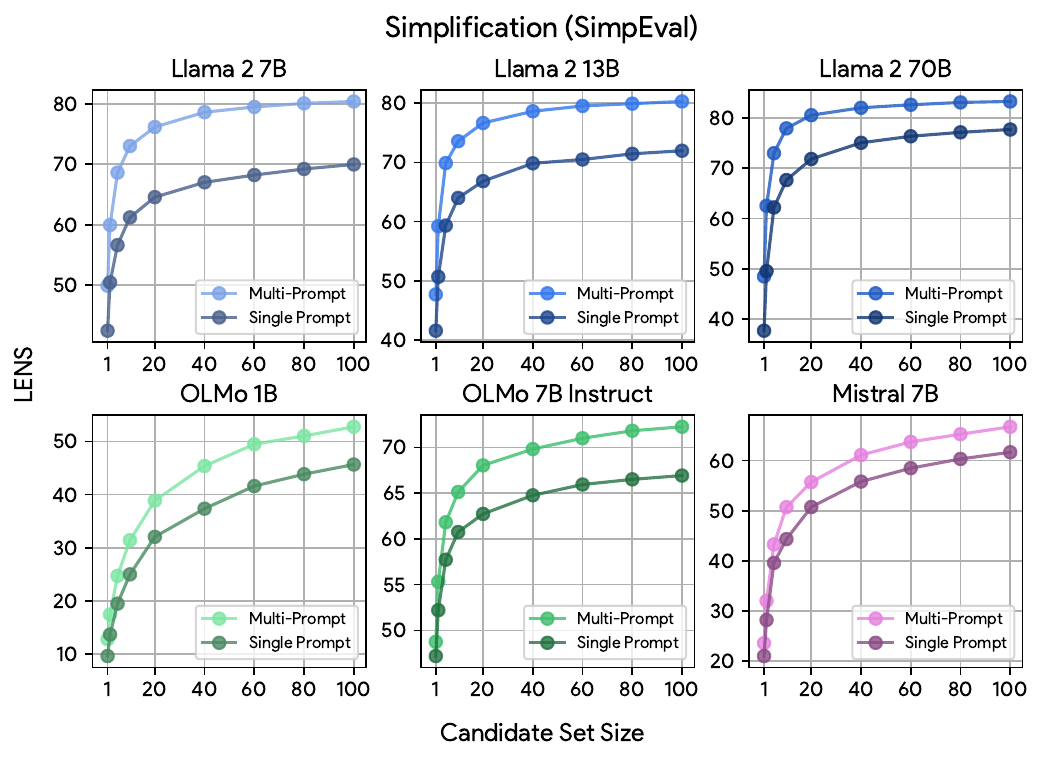}
\includegraphics[width=0.68\textwidth]{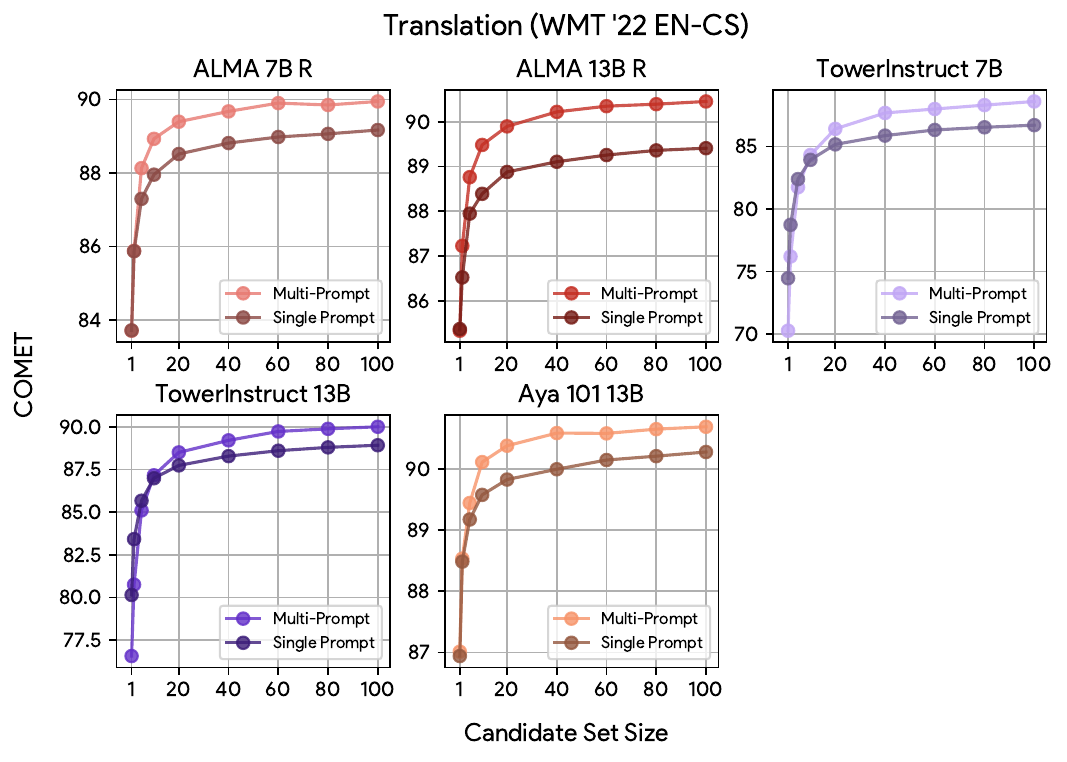}
\\[5pt]
\captionof{figure}{Results of multi-prompt MBR compared to single prompt MBR across model sizes and architectures. Multi-prompt MBR consistently improves performance across architectures and as models scale. A candidate size of 1 is equivalent to standard, single-output decoding.}
\label{fig:multi_prompt_detailed}
\end{figure*}

See Figure \ref{fig:multi_prompt_detailed} contains separated results for multi-prompt and single prompt for each model, as reported in Figure \ref{fig:multi_model} and discussed in \S\ref{sec:multi_model}.

\subsection{Detailed Cross Metric Evaluation}
Table \ref{table:cross_metric_detailed} contains the full results for the MBR experiments across metrics as discussed in \S\ref{sec:metric_eval}. While using the same metric for MBR and the final evaluation exhibits the highest improvement (see entries on the diagonal), we find that multi-prompt using any value metric universally improves performance when evaluated on any other metric. Recent neural metrics, which achieve higher correlation with human judgements, also have a higher overall performance. Note, \textsc{MetricX} scores within the range $[0,25]$ corresponding to an MQM rating, where lower is better and \textsc{Sle} scores within the range $[0,4]$ corresponding to a Newsela simplification rating, where higher is better. For clarity, we negate the \textsc{MetricX} results in Table \ref{table:cross_metric} such that all the green cells indicate a metric improvement.

\begin{table*}[t!]
\centering
\tiny
\begin{tabular}{p{0.98\linewidth}}
\toprule
\textbf{Top 10 GPT-4 Generated Text Simplification Prompts (Sorted by No. Generations Selected)} \\
\midrule

Rewrite the following sentence in a simplified manner, making sure the same meaning and message are still conveyed clearly. The simplification should be done such that it can be read and understood easily by an individual who may not have knowledge of the English language or any disabilities that limit their understanding. \\ 
\hdashrule
Please simplify the following sentence so that it is easy to understand by people with disabilities or those who are unfamiliar with English. Try to use shorter words, fewer clauses, and a simpler structure. \\ 
\hdashrule
Simplify this sentence such that a non-English speaker or a person with disabilities is able to understand the sentence. Focus on replacing complex words and structures with simpler ones, while keeping the meaning intact. You can remove unnecessary words, break up longer phrases, and generally make the text more readable. \\ 
\hdashrule
Text simplification is an important task in natural language processing for creating a simplified version of a sentence that conveys the same meaning as the original sentence but with less complex language. For this task, you will be given a sentence and asked to rewrite it using simpler words and structures so that a non-English speaker or an individual with disabilities can better understand it. Please use semantic compression to create a simplified version of the following sentence. \\ 
\hdashrule
You are an artificial intelligence designed to simplify written text. The text you are given may be complex, and your job is to rewrite it in a way that a non-english speaker or an individual with disabilities could easily understand. While you simplify the text, you should make sure it is grammatically correct and retains the original meaning of the text. \\ 
\hdashrule
You are an AI assistant tasked with creating a simpler version of a text. Text simplification can be defined as the reduction of the syntactic or lexical complexity of a text without changing its meaning. The aim of text simplification is to make the text easier to understand for a human or process by a program. Please simplify the following sentence. \\ 
\hdashrule
Rewrite this sentence in a simple and easy to understand way. Make sure to retain the meaning and ideas of the original sentence while using shorter words and sentences. \\ 
\hdashrule
Create a simpler version of the sentence below so that it can be better understood by non-English speakers or individuals with disabilities. Text simplification techniques should be used to reduce the complexity of the language while preserving the original meaning and information. \\ 
\hdashrule
You are an AI assistant that writes text simplification. Text simplification can be defined as any process that reduces the syntactic or lexical complexity of a text while attempting to preserve its meaning and information content. The aim of text simplification is to make text easier to comprehend for a human user, or process by a program. Your task is to take the following sentence and produce a simplified version that would be easier for a non-English speaker or someone with disabilities to understand. Please simplify the sentence. \\ 
\hdashrule
This prompt asks you to simplify the given sentence. In order to do so, reduce the sentence to its most basic and clear components. Remove unnecessary words, clauses, and phrases that can be inferred from the context. Use shorter, more concise words where possible.  After simplifying, the resulting sentence should still convey the same essential message. \\
\midrule
\textbf{Top 5 Randomly Sampled Few-shot Translation Instructions (Sorted by No. Generations Selected)} \\
\midrule
Anglická věta: To do this, simply access your order page, tap 'Help and support' and choose the option 'Call rider'. \\[2pt]
Česká věta: Chcete-li to provést, jednoduše přejděte na stránku objednávky, klikněte na „Nápověda a podpora“ a vyberte možnost „Zavolat jezdci“. \\[2pt]
Anglická věta: A private mass and the national anthem preceded the ceremony, which featured a portrait of De Klerk between two candles and a choir decorated with white flowers. \\[2pt]
Česká věta: Soukromá mše a státní hymna předcházely tomuto ceremoniálu, který představil portrét De Klerka mez dvěma svíčkami a sbor ozdobený bílými květy. \\[2pt]
Anglická věta: After that, we cannot offer an estimate on delivery times as it comes down to individual country's postal service and customs if outside of the EU. \\[2pt]
Česká věta: Poté nemůžeme odhadnout dobu dodání, protože záleží na poštovních a celních službách v jednotlivých zemích, pokud se nacházejí mimo EU. \\[2pt]
Anglická věta: This item is an original American comic and is in English! \\[2pt]
Česká věta: Tato položka je originální americký komiks a je v angličtině! \\[2pt]
Anglická věta: If they cannot find you they will surely call. \\[2pt]
Česká věta: Pokud vás nenajdou, určitě zavolají. \\ 
\hdashrule
Anglická věta: New Zealand's computer emergency response team was among the first to report that the flaw was being "actively exploited in the wild" just hours after it was publicly reported Thursday and a patch released. \\[2pt]
Česká věta: Tým Nového Zélandu pro reakci na počítačové ohrožení byl mezi prvními, kdo nahlásil, že tato závada se „aktivně divoce zneužívá“ jen pár hodin po tom, co byla veřejně nahlášena ve čtvrtek a byla vydána záplata. \\[2pt]
Anglická věta: Not sure, but I don't think we had any way of having them pay. \\[2pt]
Česká věta: Nejsem si jistý, ale nemyslím si, že bychom měli nějaký způsob,a by museli zaplatit. \\[2pt]
Anglická věta: Luckily, the guy was honest and rather than trying to charge the higher price, he sold me the tires for the price I had on my printout. \\[2pt]
Česká věta: Naštěstí byl ten chlapík čestný a než aby se pokoušel účtovat vyšší cenu, prodal mi pneumatiky za cenu, kterou jsem měl na mém výtisku. \\[2pt]
Anglická věta: The Cowboys just made sure Zeke and his teammates got that opportunity. \\[2pt]
Česká věta: Cowboys se právě postarali o to, aby Zeke a jeho spoluhráči tuto příležitost dostali. \\[2pt]
Anglická věta: Description Please scroll to the bottom of the listing for more pictures. \\[2pt]
Česká věta: Popis Pro více obrázků sjeďte na konec nabídky. \\ 
\hdashrule
Anglická věta: This is on a quote only basis and you need to supply us with your address for a quotation. \\[2pt]
Česká věta: Tato služba je poskytována pouze na základě cenové nabídky dle vámi poskytnuté adresy. \\[2pt]
Anglická věta: Fed up completely, she asks "Are you even going to work today?" \\[2pt]
Česká věta: Totálně znechucená se ptá: „Budeš dnes vůbec pracovat?“ \\[2pt]
Anglická věta: So there was the usual gentle chaos that attends any gathering of toddlers. \\[2pt]
Česká věta: Takže nastal obvyklý mírný chaos, který provází každé setkání batolat. \\[2pt]
Anglická věta: We currently do not have the exact information on what happened to the rider as well as to your order. \\[2pt]
Česká věta: V současné době nemáme přesné informace o tom, co se stalo s jezdcem, stejně jako s vaší objednávkou. \\[2pt]
Anglická věta: UK media reported that "thousands" were eager to raise cash for the protesters by purchasing the gray T-shirt, which depicts an empty plinth with "Bristol" written above it. \\[2pt]
Česká věta: Média ve Velké Británii hlásila, že „tisíce lidí“ nedočkavě vybírali hotovost pro protestující zakoupením šedého trička, které zobrazuje prázdný podstavec s napsaným Bristol nad ním. \\ 
\hdashrule
Anglická věta: A. No, we do not include receipts in packages unless requested. \\[2pt]
Česká věta: A. Ne, účtenku nepřikládáme, pokud to není požadováno. \\[2pt]
Anglická věta: Russia warned of 'consequences' if Ukraine attacked \\[2pt]
Česká věta: Rusko bylo varováno před “následky“, pokud napadne Ukrajinu \\[2pt]
Anglická věta: He noted that up to 90\% of all Russian investments in the Arab world are made in the UAE. \\[2pt]
Česká věta: Poznamenal, že až 90 \% ruských investicí v arabském světě jsou prováděny v SAE. \\[2pt]
Anglická věta: Many view the Softie 12 Osprey the ultimate four season synthetic fill sleeping bag available. \\[2pt]
Česká věta: Mnohými je spací pytel Softie 12 Osprey považován za nejlepší dostupný čtyřsezónní spacák se syntetickou výplní. \\[2pt]
Anglická věta: - Sign out and signing back in to your eReader. \\[2pt]
Česká věta: - Odhlaste se a přihlaste se znovu do vaší e-čtečky. \\ 
\hdashrule
Anglická věta: I told ya so.... \\[2pt]
Česká věta: Říkala jsem vám to... \\[2pt]
Anglická věta: All information about the products on our website is provided for information purposes only. \\[2pt]
Česká věta: Všechny informace o produktech na našich internetových stránkách mají pouze informativní charakter. \\[2pt]
Anglická věta: I'm in HR and have worked payroll in the past. \\[2pt]
Česká věta: Jsem na personálním oddělení a v minulosti jsem pracoval na mzdovém. \\[2pt]
Anglická věta: Years ago, I worked at a cabinet shop. \\[2pt]
Česká věta: Před lety jsem pracoval v obchodě se skříněmi. \\[2pt]
Anglická věta: De Klerk's foundation issued a posthumous video apologizing "for the pain, hurt, indignity and damage that apartheid has done" to South Africa's non-white populations. \\[2pt]
Česká věta: Fond De Klerka vydal posmrtné video omlouvající se „za bolest, zranění, ponížení a škodu, kterou apartheid udělal „jihoafrickému nebělošskému obyvatelstvu“. \\

\bottomrule
\end{tabular}
\setlength{\belowcaptionskip}{-10pt}
\caption{Prompts with highest usage for multi-prompt using the held-out split for simplification and translation.}
\label{table:prompt_selection_examples}
\end{table*}

\begin{table*}[ht!]
\centering
\small
\begin{tikzpicture}
\node(table) {
\begin{tabular}{L{62pt}C{19pt}C{19pt}C{19pt}C{19pt}C{19pt}C{19pt}}

\\[-2ex]
\multicolumn{7}{c}{\textit{Text Simplification (LLaMA 7B Chat)}} \\
\hdashrule

& 
{\hspace{-0.8cm}\rotatebox[origin=r]{-30}{{\textsc{Sari}}}} &
{\hspace{-1.3cm}\rotatebox[origin=r]{-30}{\textsc{BertScore}}} & 
{\hspace{-0.7cm}\rotatebox[origin=r]{-30}{\textsc{Lens}}} & 
{\hspace{-1.4cm}\rotatebox[origin=r]{-30}{\textsc{Lens-Salsa$^{\text{rf}}$}}} & 
{\hspace{-0.4cm}\rotatebox[origin=r]{-30}{{\textsc{Sle$^{\text{rf}}$}}}} \tabularnewline
\midrule

\textsc{Sari} & \cellcolor{blue!37} 44.33 & \cellcolor{blue!36} 92.64 & \cellcolor{blue!18} 58.73 & \cellcolor{blue!21} 72.31 & \cellcolor{blue!10} 1.42 \\
\textsc{BertScore} & \cellcolor{blue!40} 45.46 & \cellcolor{blue!40} 93.71 & \cellcolor{blue!20} 60.86 & \cellcolor{blue!20} 71.47 & \cellcolor{blue!10} 1.37 \\
\textsc{Lens} & \cellcolor{blue!26} 39.98 & \cellcolor{blue!34} 92.18 & \cellcolor{blue!40} 76.29 & \cellcolor{blue!32} 79.55 & \cellcolor{blue!21} 2.30 \\
\textsc{Lens\nobreakdash-Salsa$^{\text{rf}}$} & \cellcolor{blue!22} 38.55 & \cellcolor{blue!31} 91.29 & \cellcolor{blue!36} 73.31 & \cellcolor{blue!40} 84.59 & \cellcolor{blue!23} 2.47 \\
\textsc{Sle$^{\text{rf}}$} & \cellcolor{blue!10} 33.57 & \cellcolor{blue!10} 85.36 & \cellcolor{blue!10} 52.33 & \cellcolor{blue!10} 64.74 & \cellcolor{blue!40} 3.84 \\

\bottomrule

\rule{0pt}{2.5ex}
& \multicolumn{6}{l}{\textit{Translation (ALMA 7B)}} \\
\hdashrule

& 
% {\hspace{-1.0cm}\rotatebox[origin=r]{-30}{\textsc{Bleu}}} &
{\hspace{-1.4cm}\rotatebox[origin=r]{-30}{\textsc{BertScore}}} &
{\hspace{-1.2cm}\rotatebox[origin=r]{-30}{\textsc{Comet-22}}} &
{\hspace{-1.4cm}\rotatebox[origin=r]{-30}{\textsc{CometKiwi$^{\text{rf}}$}}} & {\hspace{-0.8cm}\rotatebox[origin=r]{-30}{\textsc{xComet}}} &
{\hspace{-0.8cm}\rotatebox[origin=r]{-30}{\textsc{MetricX}}} &
{\hspace{-1.4cm}\rotatebox[origin=r]{-30}{\textsc{MetricX-QE$^{\text{rf}}$}}} \tabularnewline
\midrule

\textsc{Bleu}                                   & \cellcolor{blue!32} 90.91 & \cellcolor{blue!19} 87.12 & \cellcolor{blue!10} 81.16 & \cellcolor{blue!16} 72.43 & \cellcolor{blue!10} 1.15 & \cellcolor{blue!10} 1.24 \\
\textsc{BertScore}                              & \cellcolor{blue!40} 91.41 & \cellcolor{blue!24} 88.11 & \cellcolor{blue!15} 82.15 & \cellcolor{blue!18} 73.59 & \cellcolor{blue!11} 1.10 & \cellcolor{blue!23} 1.15 \\
\textsc{Comet\nobreakdash-22}                   & \cellcolor{blue!26} 90.45 & \cellcolor{blue!40} 91.18 & \cellcolor{blue!40} 86.17 & \cellcolor{blue!23} 76.71 & \cellcolor{blue!21} 0.61 & \cellcolor{blue!26} 0.63 \\
\textsc{CometKiwi$^{\text{rf}}$}                & \cellcolor{blue!29} 90.67 & \cellcolor{blue!35} 90.56 & \cellcolor{blue!36} 85.64 & \cellcolor{blue!30} 81.16 & \cellcolor{blue!13} 0.51 & \cellcolor{blue!19} 0.57 \\
\textsc{xComet}                                 & \cellcolor{blue!21} 90.15 & \cellcolor{blue!34} 90.03 & \cellcolor{blue!22} 83.19 & \cellcolor{blue!40} 86.73 & \cellcolor{blue!17} 0.70 & \cellcolor{blue!21} 0.79 \\
\textsc{MetricX}                                & \cellcolor{blue!10} 89.35 & \cellcolor{blue!29} 89.07 & \cellcolor{blue!14} 82.00 & \cellcolor{blue!10} 69.26 & \cellcolor{blue!37} 0.47 & \cellcolor{blue!37} 0.69 \\
\textsc{MetricX\nobreakdash-QE$^{\text{rf}}$}   & \cellcolor{blue!13} 89.58 & \cellcolor{blue!30} 89.29 & \cellcolor{blue!26} 83.93 & \cellcolor{blue!10} 68.78 & \cellcolor{blue!40} 0.43 & \cellcolor{blue!40} 0.25 \\

\bottomrule
\end{tabular}
};
\coordinate (nw) at (table.north west);
\coordinate (ne) at (table.north east);
\coordinate (sw) at (table.south west);
\draw[thick, -latex] ($(nw) + (0, 0.25em)$) -- ($(ne) + (0, 0.25em)$)
node[midway, fill=white] {Evaluation Metric};
\draw[thick, -latex] ($(nw) + (-0.5em, 0)$) -- ($(sw) + (-0.5em, 0)$)
node[midway, fill=white, rotate=90] {MBR Utility Metric};
\end{tikzpicture}
\begin{tikzpicture}
\node (table) {
\begin{tabular}{L{62pt}C{19pt}C{19pt}C{19pt}C{19pt}C{19pt}C{19pt}}

\\[-2ex]
\multicolumn{7}{c}{\textit{Text Simplification (LLaMA 7B Chat)}} \\
\hdashrule

& 
{\hspace{-0.8cm}\rotatebox[origin=r]{-30}{{\textsc{Sari}}}} &
{\hspace{-1.3cm}\rotatebox[origin=r]{-30}{\textsc{BertScore}}} & 
{\hspace{-0.7cm}\rotatebox[origin=r]{-30}{\textsc{Lens}}} & 
{\hspace{-1.4cm}\rotatebox[origin=r]{-30}{\textsc{Lens-Salsa$^{\text{rf}}$}}} & 
% {\hspace{-0.4cm}\rotatebox[origin=r]{-30}{{\textsc{Bets}}}} & 
{\hspace{-0.4cm}\rotatebox[origin=r]{-30}{{\textsc{Sle$^{\text{rf}}$}}}} \tabularnewline
\midrule

\textsc{Sari} & \cellcolor{red!35} 43.25 & \cellcolor{red!27} 91.58 & \cellcolor{red!12} 51.49 & \cellcolor{red!10} 67.97 & \cellcolor{red!11} 1.04 \\
\textsc{BertScore} & \cellcolor{red!40} 44.02 & \cellcolor{red!40} 92.62 & \cellcolor{red!16} 54.68 & \cellcolor{red!10} 68.36 & \cellcolor{red!10} 0.92 \\
\textsc{Lens} & \cellcolor{red!20} 40.64 & \cellcolor{red!35} 92.24 & \cellcolor{red!40} 70.51 & \cellcolor{red!27} 74.86 & \cellcolor{red!19} 1.49 \\
\textsc{Lens\nobreakdash-Salsa$^{\text{rf}}$} & \cellcolor{red!13} 39.38 & \cellcolor{red!20} 90.94 & \cellcolor{red!32} 65.21 & \cellcolor{red!40} 79.93 & \cellcolor{red!19} 1.51 \\
\textsc{Sle$^{\text{rf}}$} & \cellcolor{red!10} 38.82 & \cellcolor{red!10} 90.07 & \cellcolor{red!10} 49.94 & \cellcolor{red!13} 69.26 & \cellcolor{red!40} 2.79 \\

\bottomrule

\rule{0pt}{2.5ex}
& \multicolumn{6}{l}{\textit{Translation (ALMA 7B)}} \\
\hdashrule

& 
% {\hspace{-1.0cm}\rotatebox[origin=r]{-30}{\textsc{Bleu}}} &
{\hspace{-1.4cm}\rotatebox[origin=r]{-30}{\textsc{BertScore}}} &
{\hspace{-1.2cm}\rotatebox[origin=r]{-30}{\textsc{Comet-22}}} &
{\hspace{-1.4cm}\rotatebox[origin=r]{-30}{\textsc{CometKiwi$^{\text{rf}}$}}} & {\hspace{-0.8cm}\rotatebox[origin=r]{-30}{\textsc{xComet}}} &
{\hspace{-0.8cm}\rotatebox[origin=r]{-30}{\textsc{MetricX}}} &
{\hspace{-1.4cm}\rotatebox[origin=r]{-30}{\textsc{MetricX-QE$^{\text{rf}}$}}} \tabularnewline
\midrule

\textsc{Bleu}                                   & \cellcolor{red!34} 90.57 & \cellcolor{red!10} 86.65 & \cellcolor{red!10} 80.49 & \cellcolor{red!23} 72.57 & \cellcolor{red!10} 1.20 & \cellcolor{red!10} 1.35 \\
\textsc{BertScore}                              & \cellcolor{red!40} 90.90 & \cellcolor{red!10} 86.52 & \cellcolor{red!10} 80.48 & \cellcolor{red!21} 71.10 & \cellcolor{red!11} 1.31 & \cellcolor{red!19} 1.44 \\
\textsc{Comet\nobreakdash-22}                   & \cellcolor{red!21} 89.74 & \cellcolor{red!36} 90.28 & \cellcolor{red!28} 84.44 & \cellcolor{red!24} 73.42 & \cellcolor{red!20} 0.74 & \cellcolor{red!24} 0.81 \\
\textsc{CometKiwi$^{\text{rf}}$}                & \cellcolor{red!23} 89.87 & \cellcolor{red!34} 89.53 & \cellcolor{red!40} 84.58 & \cellcolor{red!32} 78.29 & \cellcolor{red!12} 0.58 & \cellcolor{red!19} 0.65 \\
\textsc{xComet}                                 & \cellcolor{red!26} 90.01 & \cellcolor{red!28} 89.18 & \cellcolor{red!18} 82.35 & \cellcolor{red!40} 83.39 & \cellcolor{red!18} 0.79 & \cellcolor{red!23} 0.83 \\
\textsc{MetricX}                                & \cellcolor{red!10} 88.99 & \cellcolor{red!22} 88.26 & \cellcolor{red!15} 81.63 & \cellcolor{red!12} 65.32 & \cellcolor{red!40} 0.54 & \cellcolor{red!40} 0.66 \\
\textsc{MetricX\nobreakdash-QE$^{\text{rf}}$}   & \cellcolor{red!10} 88.98 & \cellcolor{red!17} 87.61 & \cellcolor{red!16} 81.82 & \cellcolor{red!10} 63.47 & \cellcolor{red!35} 0.50 & \cellcolor{red!37} 0.27 \\

\bottomrule
\end{tabular}
};

\coordinate (nw) at (table.north west);
\coordinate (ne) at (table.north east);
\draw[thick, -latex] ($(nw) + (0, 0.25em)$) -- ($(ne) + (0, 0.25em)$)
node[midway, fill=white] {Evaluation Metric};
\end{tikzpicture}
\caption{\textcolor{blue}{Multi-prompt} and \textcolor{red}{single prompt} performance across metrics. \textsc{rf $=$} Reference-free reranker.}
\label{table:cross_metric_detailed}
\end{table*}

\end{document}